\documentclass[journal]{IEEEtran}

\usepackage{booktabs}
\usepackage{graphicx}
\usepackage{amsmath}
\usepackage{amssymb}
\usepackage[table]{xcolor}
\usepackage{multirow}
\usepackage{tabularx}
\usepackage{algorithm}
\usepackage{algpseudocode}

\usepackage{booktabs}
\usepackage{tabularx}
\usepackage{array}
\usepackage{amssymb}

\usepackage{amsmath,amssymb}
\usepackage{graphicx}
\usepackage{booktabs}
\usepackage{multirow}
\usepackage{tabularx}
\usepackage{array}
\usepackage[table]{xcolor}
\usepackage{cite}

\usepackage{tabularray}
\UseTblrLibrary{booktabs}

\definecolor{rdoLink}{RGB}{0,90,140}
\definecolor{rdoBest}{gray}{0.93}
\newcommand{\best}[1]{\textbf{#1}}
\newcolumntype{Y}{>{\centering\arraybackslash}X}

\usepackage[
  colorlinks=true,
  linkcolor=rdoLink,
  citecolor=rdoLink,
  urlcolor=rdoLink,
  pdfauthor={}
]{hyperref}

\begin{document}

\title{RaMOD: Moving-Object Detection from Raw mmWave Radar Cubes under Ego Motion}

\author{%
\resizebox{\textwidth}{!}{%
Yuan~Fang$^{1}$, Long~Zhuang$^{2}$,
Lingzi~Xing$^{3}$, Xijia~Wei$^{1}$,
Jingqi~Jiang$^{4}$, Fangzhan~Shi$^{1}$,
Simon~Julier$^{1}$, and Kevin~Chetty$^{1}$%
}%
\thanks{$^{1}$University College London, London, United Kingdom.}%
\thanks{$^{2}$Southeast University, Nanjing, China, and
Nanyang Technological University, Singapore.
(Corresponding author.)}%
\thanks{$^{3}$University of Liverpool, Liverpool, United Kingdom.}%
\thanks{$^{4}$Imperial College London, London, United Kingdom.}%
}

\maketitle

\begin{abstract}
Millimeter-wave (mmWave) radar provides weather-robust sensing and
direct Doppler measurements, yet automotive perception pipelines often
reduce its output to sparse point clouds or two-dimensional
projections, leaving raw range--azimuth--Doppler (RAD) cubes
underutilized. Ego motion induces Doppler responses from world-static
structures, while Doppler wrapping aliases radial velocities,
complicating moving-object detection and velocity estimation. We
present RaMOD (Radar-based Moving-Object Detection), a real-time
framework for moving-object detection and velocity estimation from
raw RAD cubes acquired by a vehicle-mounted single-chip radar. RaMOD
explicitly estimates frame-wise ego motion and uses the estimated
radar translational velocity to predict the Doppler region of
world-static structures and suppress returns within it. It temporally
encodes the suppressed cubes into multi-scale features to predict
RAD-space boxes with class labels. For each detected object, it
predicts radial and forward velocities under raw-Doppler and
Doppler-unwrapped hypotheses, together with Doppler-wrapping
probabilities. On our BRIDGE-Liv dataset, RaMOD outperforms all
compared raw-radar methods across RAD-cube, RA, and RD detection
settings, achieving $80.8\%$ cube mAP at IoU $0.30$, $15.7$ percentage
points above the strongest RAD-cube baseline. Velocity MAE is
$1.128~\mathrm{m/s}$, with $99.87\%$ Doppler-wrapping accuracy.
\end{abstract}


\IEEEpeerreviewmaketitle

\section{Introduction}

\IEEEPARstart{M}{oving} object detection is a fundamental perception task for assisted and autonomous driving, where independently moving road users must be accurately localized, classified, and assigned reliable motion estimates. In current autonomous driving systems, optical sensors, especially cameras and LiDARs, play a decisive role in perception. In conventional modular pipelines, cameras provide rich semantic information for object detection and segmentation~\cite{li2022bevformer}, while LiDAR provides accurate range and 3D geometric measurements for object detection~\cite{lang2019pointpillars}. Recent end-to-end driving frameworks further learn driving decisions from sensor observations~\cite{Hu_2023_CVPR}, while emerging vision-language-action (VLA) systems incorporate language guidance into driving~\cite{zhou2025opendrivevla}. However, object motion in conventional camera- and LiDAR-based pipelines is typically inferred through temporal association, optical flow, scene flow, or frame-to-frame geometric changes, which can be sensitive to occlusion and unreliable correspondences~\cite{teed2020raft,liu2019flownet3d}. Moreover, cameras can degrade under poor illumination, glare, contamination, and motion blur, while LiDAR measurements can be affected by rain, fog, and snow~\cite{Bijelic_2020_CVPR}, as well as smoke~\cite{millimap}.

Millimeter-wave radar has therefore become an important complementary sensor in modern vehicles. Compared with optical sensors, mmWave radar uses longer-wavelength electromagnetic signals and is generally more robust under poor illumination and adverse weather conditions. More importantly, radar directly measures radial velocity through Doppler. Advances in semiconductor fabrication and antenna-in-package technologies have also enabled compact, low-power, and low-cost single-chip mmWave radars, making large-scale deployment practical in intelligent vehicles and autonomous systems \cite{1843aop}. Automotive radars have consequently been widely adopted in advanced driver-assistance systems (ADAS).

However, automotive radar perception still commonly relies on processed point clouds and the clusters derived from them (Fig.~\ref{fig:overview}(c)). Point clouds are typically generated through peak selection and sparsification, while clustering, data association, and tracking further produce object-level estimates. The resulting detections and tracks provide compact position and velocity information for conventional ADAS functions such as adaptive cruise control, blind-spot detection, collision warning, and automatic emergency braking (AEB)~\cite{patole2017automotive}. Nevertheless, peak-based sparsification may discard weak responses and only partially preserve the extended object support and reflection structures present in the full radar measurement tensor (Fig.~\ref{fig:overview}(b)).

For next-generation autonomous driving, mmWave radar has the potential to support high-level perception through its spatial measurements and directly measured Doppler information. Raw range--azimuth--Doppler (RAD) cubes preserve these measurement distributions before point-cloud sparsification, providing richer evidence for learning-based object detection, classification, and velocity estimation. Among these capabilities, identifying independently moving road users through spatial and Doppler evidence is particularly important for interaction, motion prediction, and collision-risk assessment.

However, identifying independently moving road users becomes more
challenging when the radar itself is moving. World-static structures,
such as buildings and parked vehicles, also produce Doppler shifts
under ego motion, so non-zero Doppler alone does not indicate
independent object motion. Their responses may overlap with those of
moving objects. Doppler wrapping introduces further ambiguity: when
the relative radial velocity falls outside the unambiguous velocity
interval, the return is mapped to an aliased Doppler bin, potentially
increasing static--dynamic overlap. The extent of this overlap depends
on radar Doppler coverage and the relative motions in the scene.
The world-static Doppler field and the resulting ambiguity after
wrapping are analyzed in
Secs.~\ref{sec:world_static_doppler_field} and
\ref{sec:doppler_wrapping_static_dynamic_ambiguity}.

Motivated by these observations, we propose RaMOD, a real-time
framework for moving-object detection from raw RAD cubes acquired
by a single-chip radar on a moving vehicle. It supports both
single-frame and short-sequence inputs. Since
world-static Doppler depends on radar motion and viewing direction,
RaMOD estimates frame-wise ego motion as an explicit output and uses
the estimated radar translational velocity to predict the
static-consistent Doppler region. Suppressing returns within this
region reduces ego-motion-induced background responses before
detection. The suppressed radar representation is encoded into
multi-scale features for RAD-space object localization and
classification. RaMOD also predicts radial and
object-level forward velocities under raw-Doppler and
Doppler-unwrapped hypotheses, together with Doppler-wrapping
probabilities. Experiments on BRIDGE-Liv, which contains moving-ego
urban and highway sequences, show that RaMOD outperforms all compared 
raw-radar methods across RAD-cube, range--azimuth (RA), and range--Doppler (RD)
detection settings.

The main contributions of this paper are summarized as follows:
\begin{itemize}
\item We propose RaMOD, a real-time moving-object detection and motion-estimation framework that directly processes raw RAD cubes from a single-chip radar under ego motion. It retains Doppler as an explicit detection dimension and combines multi-scale cube encoding, cross-scale feature refinement, and cross-view RA/RD auxiliary supervision for RAD-space object localization and classification.

\item We develop an ego-motion-guided Doppler suppression mechanism. Frame-wise ego motion is estimated directly from unmasked RAD cubes as an explicit output. Its translational component is combined with the world-static Doppler model to construct an internal cube-level mask that suppresses returns within the predicted static-consistent Doppler region before detection.

\item We design object-level velocity heads that predict radial and forward velocities under raw-Doppler and Doppler-unwrapped hypotheses, together with Doppler-wrapping probabilities. These probabilities guide selection between the two velocity pairs, associating each detected object with a velocity estimate that accounts for Doppler ambiguity.
\end{itemize}

\section{Related Work}

\subsection{Camera- and LiDAR-based Automotive Perception}

Cameras and LiDARs are the dominant sensing modalities for
high-level automotive perception. Camera-based methods have progressed
from two-stage and real-time object detectors to transformer-based 3D
detection, BEV representation, and occupancy prediction
~\cite{girshick2014rcnn,redmon2016yolo,carion2020detr,
wang2022detr3d,li2022bevformer,wei2023surroundocc}. LiDAR-based
methods use accurate range and 3D geometry for multi-class bounding-box
detection from point clouds
~\cite{zhou2018voxelnet,lang2019pointpillars,shi2020pvrcnn,
yin2021centerpoint}.

Motion perception with these sensors generally relies on temporal
visual or geometric changes, including tracking-by-detection, optical
flow, scene flow, temporal BEV fusion, and multi-scan association
~\cite{zhang2022bytetrack,teed2020raft,liu2019flownet3d,
li2022bevformer}. LiDAR moving-object segmentation further distinguishes
moving points from the static scene
~\cite{chen2021movingobjectsegmentation,wu2024mdetector}. This
distinction is important because a road user may be movable without
being currently in motion. In contrast, mmWave radar directly measures
radial velocity through Doppler and remains robust under poor
illumination and adverse weather, making it particularly suitable for
motion-aware automotive perception.

\subsection{Radar Point-Cloud Perception and Ego-Motion Estimation}

Radar point clouds are widely used because they provide compact target
measurements after peak detection and thresholding. Early learning-based
methods perform object classification and 2D or 3D detection from sparse
radar points
~\cite{danzer2019pointnet,scheiner2021radarpc,palffy2022vod}.
Subsequent methods introduce radar-specific feature aggregation to
address sparsity and noise. RPFA-Net employs pillar feature attention,
SMURF combines density features with multiple representations, and
RadarPillars decomposes radial velocity before pillar-based detection
~\cite{xu2021rpfa,liu2023smurf,musiat2024radarpillars}.

Doppler is useful in these detectors but is generally treated as an
auxiliary per-point feature. For example, the View-of-Delft study
reports that removing Doppler reduces radar detection performance from
$38.0$ to $29.1$ mAP~\cite{palffy2022vod}. Nevertheless, the resulting
networks typically predict boxes, classes, and confidence scores rather
than object-associated radial velocity, Doppler-wrapping states, or
unwrapped velocity. Moreover, peak-based point-cloud generation may discard weak
responses and only partially preserve extended object support,
motivating lower-level representations that retain dense spatial
and Doppler structure.

Radar-based ego-motion estimation constitutes a related research
direction. Classical automotive methods estimate instantaneous ego velocity from
the azimuth-dependent Doppler measurements of world-static radar
detections and infer yaw rate under a vehicle-motion model
~\cite{kellner2013instantaneous}. Subsequent approaches combine Doppler
constraints with spatial registration
~\cite{rapp2017probabilistic}, learn to identify static inliers for
robust instantaneous estimation
~\cite{zhu2023deepego}, or derive translational and rotational odometry
from Doppler--azimuth and range--azimuth heatmaps in radar-only SLAM
systems~\cite{sie2024radarize}. RaMOD incorporates frame-wise ego-motion
estimation as an explicit network output and uses the predicted
translational velocity to construct its internal static-consistent
Doppler suppression region.

\subsection{Low-Level and Raw Radar Perception}

Projection-based heatmap methods perform object detection or semantic
segmentation in two-dimensional radar views. On the stationary-radar
CARRADA benchmark~\cite{ouaknine2021carrada}, multi-view convolutional,
attention, and state-space models learn radar-view semantics~\cite{ouaknine2021mvnet,zhang2023peakconv,li2024adapkc,
chen2026marss}, while RD detectors predict bounding boxes~\cite{decourt2022darod}.
On vehicle-mounted datasets such as UWCR and CRUW, temporal models
aggregate consecutive RA heatmaps or multiple radar views through
spatiotemporal convolutions, recurrent memory, or attention~\cite{gao2021rampcnn,wang2021rodnet,jiang2023trodnet,
decourt2024record,dalbah2023radarformer}. Their outputs principally
describe class-labeled object locations or semantic regions in
radar views. A single-frame hybrid alternative combines extracted
radar targets with local RAD crops for classification and subsequent
object clustering~\cite{palffy2020rtcnet}.

RAD-space detectors instead predict boxes spanning range, azimuth,
and Doppler. Existing designs include anchor-based convolutional
detection~\cite{zhang2021raddet}, transformer-based encoding with
anchor-free heads~\cite{cheng2025transrad}, and NMS-free set
prediction~\cite{zhang2026ptf}. These outputs retain Doppler
localization, but the RADDet benchmark used in these studies was
collected with a stationary radar and does not capture the
ego-motion-induced static Doppler field encountered by a moving sensor.

Another line of work learns spatial features directly from RD
spectra or ADC signals. On DDM data in RADIal,
learned signal transformations, transformer encoders, and sparse
feature extraction support vehicle detection and free-space
segmentation~\cite{rebut2022radial,yang2023adcnet,
giroux2023tfftradnet,wu2024sparseradnet}. The RADIal detection
benchmark considers a single vehicle class. Complementary
ADC--heatmap fusion enriches detection features on RADDet~\cite{zhuang2025rawadc}.

Radar representations also support detection in physical 3D
space~\cite{fang2025cubedn} and human
pose estimation~\cite{wei2026maepose}. In automotive settings, volumetric
radar features support 3D bounding-box detection~\cite{paek2022kradar,paek2023enhancedkradar,kong2025rtnhplus}
and tracking~\cite{cheng2023centerradarnet}, while high-resolution
RA heatmaps support oriented vehicle detection~\cite{madani2022radatron}.
Temporal RAD tensors augmented with elevation-index information
also enable binary 3D occupancy prediction~\cite{roldan2024radelft}.
A later RaDelft-based method predicts voxel-level semantic
labels~\cite{sun2025automatic}, while RadarOcc predicts foreground,
background, and free-space occupancy on K-Radar~\cite{ding2024radarocc}.
These methods focus on spatial geometry, semantics, and tracking rather
than object-level Doppler-wrapping decisions and unwrapped
radial-velocity outputs.

Object-level velocity estimation and signal-level Doppler unfolding
have also been investigated. Major et al. combine projected radar
representations with temporal modeling to estimate an ego-relative
velocity vector for each detected vehicle~\cite{major2019vehicle}.
For vehicle-mounted TDM-MIMO radar, Zheng et al.'s TDRadarNet
pipeline combines learned Doppler unfolding with phase-migration
compensation before temporal BEV vehicle detection~\cite{zheng2023tdradarnet}.
Its unfolding stage resolves ambiguity for image formation rather
than assigning a wrapping state to each detected object. In comparison,
RaMOD combines ego-motion-guided cube suppression, RAD-space detection,
and object-associated velocity estimation with explicit wrapping
predictions.

\begin{table*}[!t]
\centering
\caption{Comparison of representative low-level automotive radar
perception methods.}
\label{tab:capability_comparison}
\scriptsize

\begin{tblr}{
width=\textwidth,
colspec={
Q[c,m,wd=0.050\textwidth]
Q[c,m,wd=0.075\textwidth]
X[l,m]
Q[c,m,wd=0.090\textwidth]
Q[c,m,wd=0.062\textwidth]
Q[c,m,wd=0.075\textwidth]
Q[c,m,wd=0.062\textwidth]
Q[c,m,wd=0.065\textwidth]
Q[c,m,wd=0.062\textwidth]
Q[c,m,wd=0.052\textwidth]
},
colsep=1.5pt,
rowsep=1.8pt,
row{1}={font=\bfseries},
cell{2}{1}={r=2}{c,m},
cell{4}{1}={r=9}{c,m},
}
\toprule
{Radar\\State}
& Dataset
& Representative Methods
& Input
& Location
& {\mbox{Multi-class}\\Labels}
& {Radial\\Velocity}
& {Wrap/\\Unwrap}
& {Object\\Velocity}
& {Ego\\Motion}
\\
\midrule

Fixed
& CARRADA
& MV-Net, TMVA-Net, PeakConv, AdaPKC, MARSS, DAROD~\cite{ouaknine2021mvnet,
zhang2023peakconv,li2024adapkc,chen2026marss,decourt2022darod}
& RA/RD/AD
& \checkmark & \checkmark & -- & -- & -- & --
\\

& RADDet
& RADDet, TransRAD, Mamba-RODNet, PTF~\cite{zhang2021raddet,
cheng2025transrad,zhuang2025rawadc,zhang2026ptf}
& RAD/ADC
& \checkmark & \checkmark & -- & -- & -- & --
\\
\midrule

Moving
& Private
& Major et al.~\cite{major2019vehicle}
& RAD proj.
& \checkmark & -- & -- & -- & \checkmark$^{\dagger}$ & --
\\

& Private
& RTCnet~\cite{palffy2020rtcnet}
& Targets+RAD
& Clust. & \checkmark & -- & -- & -- & --
\\

& UWCR
& RAMP-CNN~\cite{gao2021rampcnn}
& RVA
& \checkmark & \checkmark & -- & -- & -- & --
\\

& CRUW
& RODNet, T-RODNet, RECORD, RadarFormer~\cite{wang2021rodnet,
jiang2023trodnet,decourt2024record,dalbah2023radarformer}
& RA
& \checkmark & \checkmark & -- & -- & -- & --
\\

& RADIal
& FFT-RadNet, ADCNet, T-FFTRadNet, SparseRadNet~\cite{rebut2022radial,
yang2023adcnet,giroux2023tfftradnet,wu2024sparseradnet}
& ADC/RD
& \checkmark & -- & -- & -- & -- & --
\\

& K-Radar
& RTNH, Enhanced K-Radar, RTNH+, CenterRadarNet~\cite{paek2022kradar,
paek2023enhancedkradar,kong2025rtnhplus,cheng2023centerradarnet}
& RDAE
& \checkmark & \checkmark & -- & -- & -- & --
\\

& Radatron
& Radatron~\cite{madani2022radatron}
& RA
& \checkmark & -- & -- & -- & -- & --
\\

& Private
& TDRadarNet, TDRadarNet+~\cite{zheng2023tdradarnet}
& BEV spectra
& \checkmark & -- & -- & --$^{\ast}$ & -- & --
\\
\cmidrule{2-10}

& BRIDGE-Liv
& \textbf{RaMOD (ours)}
& RAD
& \checkmark & \checkmark & \checkmark & \checkmark
& \checkmark$^{\ddagger}$ & \checkmark
\\
\bottomrule
\end{tblr}

\vspace{1mm}
\begin{minipage}{\textwidth}
\scriptsize
Radar state refers to data collection.
$\checkmark$/--: provided/not provided by the method.
Multi-class denotes multiple foreground categories.
Velocity and wrapping refer to object-level outputs, not input
Doppler or box coordinates; external ego-motion inputs do not count.
Clust.: target clustering.
RDAE: range--Doppler--azimuth--elevation data, with Doppler reduced
before detection in some methods. $^{\dagger}$Ego-relative velocity magnitude and direction.
$^{\ddagger}$Ground-referenced forward velocity assuming road-aligned motion.
$^{\ast}$Doppler unfolding is used to form radar images, not predicted
for each detected object.
\end{minipage}
\vspace{-3mm}
\end{table*}

Among these methods, representative approaches for RA detection
(including methods originally based on confidence maps)~\cite{wang2021rodnet,
jiang2023trodnet,dalbah2023radarformer,rebut2022radial,
giroux2023tfftradnet}, RA/RD multi-view perception~\cite{ouaknine2021mvnet,
zhang2023peakconv,li2024adapkc}, and RAD-cube detection~\cite{zhang2021raddet,zhuang2025rawadc}
are evaluated as baselines on BRIDGE-Liv
(Sec.~\ref{sec:sota_comparison}). Table~\ref{tab:capability_comparison} compares the prediction
outputs of representative methods, including RaMOD. It also
indicates whether their datasets were collected with stationary
or moving radars.

\section{mmWave Radar Cube Generation and Doppler-Aware Object Representation}

\subsection{FMCW Measurement Principles and Radar Cube Generation}

\begin{figure*}[t]
  \centering
  \includegraphics[width=1.0\textwidth]{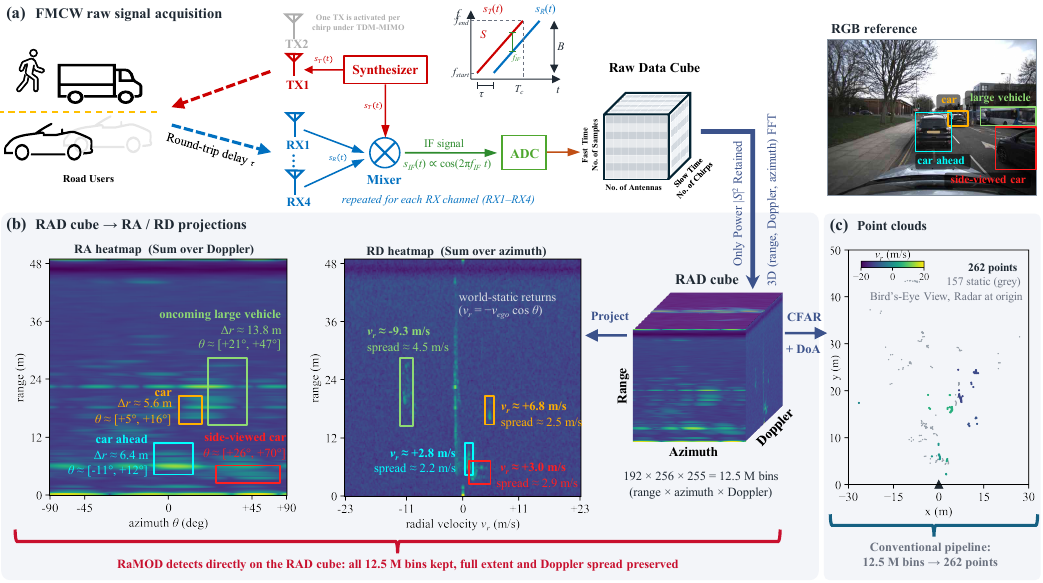}
  \caption{(a) Signal chain of the 2-TX, 4-RX TDM-MIMO FMCW radar. Each
  received chirp $s_R(t)$ is mixed with the transmitted chirp $s_T(t)$
  to produce the IF signal $s_{\mathrm{IF}}(t)$, which the ADC samples
  into the raw data cube of ADC samples $\times$ chirps $\times$
  antennas. (b) A 3D FFT converts it into the range--azimuth--Doppler
  (RAD) cube, on which RaMOD detects directly. Its RA and RD projections
  are shown only for visualization, revealing four road users at
  different azimuths with distinct radial velocities and Doppler
  spreads. (c) The conventional pipeline applies CFAR and
  direction-of-arrival (DoA) estimation to the same cube, compressing
  12.5\,M bins into only 262 points (157 static). Object extent and
  class are hard to recover from such points, and further clustering
  and tracking are still required.}
  \label{fig:overview}
  \vspace{-4mm}
\end{figure*}

FMCW mmWave radar transmits linearly frequency-modulated chirps and
mixes received echoes with the transmitted signal to produce the
intermediate-frequency (IF) signal $s_{\mathrm{IF}}(t)$. ADC sampling
forms a raw data cube organized along fast time, slow time, and
antenna channels (Fig.~\ref{fig:overview}(a)).

The fast-time beat frequency primarily encodes range through
$f_{\mathrm{IF}}=2SR/c$, where $R$ is the target range, $S$ is the
chirp slope, and $c$ is the speed of light. The range resolution is
$\Delta R=c/(2B)$, where $B$ is the chirp bandwidth.

Doppler information is encoded by the phase evolution between
consecutive chirps from the same transmitter. For phase change
$\Delta\phi_{\mathrm{IF}}$ and effective sampling interval
$T_{\mathrm{eff}}$, the Doppler frequency is
$f_D=\Delta\phi_{\mathrm{IF}}/(2\pi T_{\mathrm{eff}})$, with
$T_{\mathrm{eff}}$ defined in Sec.~\ref{sec:doppler_coverage_design}.
The corresponding radial velocity is $v_r=\lambda f_D/2$,
where $\lambda$ is the radar wavelength.

After Doppler FFT, the signed radial-velocity axis contains
$N_{\mathrm{D}}$ bins over $[-v_{\max},v_{\max})$, with spacing
$\Delta v_{\mathrm{D}}=2v_{\max}/N_{\mathrm{D}}$, where $v_{\max}$
is the maximum unambiguous radial velocity. Throughout this paper,
each Doppler bin is represented by its bin-center velocity
$v_{\mathrm{D}}$ rather than its index.

The angle of arrival is estimated from the phase differences
$\Delta\phi_{\theta}=2\pi d\sin\theta/\lambda$ across antenna elements,
where $d$ is the antenna spacing and $\theta$ is the azimuth angle
relative to the radar boresight. The azimuth axis samples the usable
field of view $\Theta_{\mathrm{FoV}}$. Its sampling grid depends on
the antenna geometry and FFT or beamforming configuration, whereas
angular resolution is mainly determined by the effective array
aperture.

Range, Doppler, and angle FFTs yield the RAD cube, with each cell
representing one range--azimuth--Doppler bin. Only the squared
magnitude (power) of the spectrum is retained
(Fig.~\ref{fig:overview}(b)).

\subsection{Radar Cube Representations and Object Measurements}

\subsubsection{RA, RD, and RAD Representations}
\label{sec:ra_rd_rad_representations}
A RAD cube can be projected into two-dimensional views by summing
along one axis (Fig.~\ref{fig:overview}(b)): the RA projection sums
over Doppler and shows object locations in the polar plane, whereas
the RD projection sums over azimuth and exposes Doppler structure.
Each projection is $\log_{10}$-compressed and, for visualization,
normalized independently, so color indicates relative intensity only.
These projections are convenient to inspect but discard the
azimuth--Doppler association of individual returns, which only the
full RAD cube preserves. In this work, RA and RD serve visualization,
annotation, cross-view auxiliary supervision, and proposal guidance,
while detection is performed directly in the RAD cube.

\subsubsection{Object Measurement Extent in Radar Cubes}
\label{sec:extended_object_measurements}

Road users should not be treated as ideal point targets in radar
cubes. A road user can produce multiple scattering centers and occupy
a finite measurement extent in range, azimuth, and Doppler. This
extent depends on both the physical size of the object and its viewing
geometry.

As illustrated in Fig.~\ref{fig:overview}(b), large vehicles such as
buses and trucks generally have longer bodies and may therefore occupy
more range bins. The front-viewed large vehicle highlighted in green
shows a substantial range extent, whereas passenger cars typically
occupy a more compact range region. Viewing geometry also has a stronger effect on the azimuth extent. For a front- or rear-viewed vehicle, its scattering centers usually have similar azimuth angles and therefore remain relatively compact in the angular dimension. In contrast, a nearby or side-viewed vehicle can span a much wider azimuth interval because different parts of the vehicle are observed from noticeably different line-of-sight directions. Consequently, the same road user may appear as an extended region in RA rather than as a single point.

The angular extent of an independently moving object can also produce
a Doppler spread. To isolate this effect from ego motion, consider a static radar platform, i.e.,
$\mathbf{v}_{\mathrm{ego}}=0$. For a moving object with velocity
$\mathbf{v}_{\mathrm{dyn}}$, the radial velocity of a scattering center
observed at azimuth angle $\theta$ is
\begin{equation}
v_r^{\mathrm{dyn}}(\theta)
=
\mathbf{v}_{\mathrm{dyn}}^{\top}
\mathbf{u}(\theta),
\end{equation}
where $\mathbf{u}(\theta)$ is the line-of-sight unit vector from the
radar to the scattering center. Here, \emph{dynamic} refers to
objects moving relative to the world, and the label $\mathrm{dyn}$
denotes their associated quantities.

If the object spans an angular interval $[\theta_1,\theta_2]$, its
scattering centers are observed from different line-of-sight
directions and may therefore have different radial-velocity
projections. When the radial-velocity projection only increases or
only decreases over $[\theta_1,\theta_2]$, the corresponding Doppler
spread can be approximated as
\begin{equation}
\Delta v_r^{\mathrm{dyn}}
\approx
\left|
\mathbf{v}_{\mathrm{dyn}}^{\top}
\left(
\mathbf{u}(\theta_1)
-
\mathbf{u}(\theta_2)
\right)
\right|.
\end{equation}

If the object motion is also approximately along the radar boresight
direction, this expression simplifies to
\begin{equation}
\Delta v_r^{\mathrm{dyn}}
\approx
\left\|
\mathbf{v}_{\mathrm{dyn}}
\right\|
\left|
\cos\theta_1-\cos\theta_2
\right|.
\end{equation}
Therefore, even a single moving vehicle does not necessarily
correspond to a single Doppler peak. As shown in
Fig.~\ref{fig:overview}(b), the front-viewed vehicle remains relatively
compact in azimuth and produces a narrow Doppler distribution. In
contrast, the side-viewed vehicle highlighted in red spans a much
wider angular interval and occupies a broader Doppler region because
its scattering centers have substantially different radial-velocity
projections. For motion along the radar boresight, scattering centers
observed near $|\theta|=90^\circ$ may have near-zero radial velocities.

More generally, differences in physical size, scattering-center
distribution, and motion structure cause large vehicles, passenger
cars, pedestrians, and two-wheelers to exhibit characteristic
range--azimuth extents and Doppler patterns. These class- and
viewpoint-dependent signatures provide important cues for radar-based
detection and classification, motivating cube-level bounding boxes
rather than center-only labels. In RaMOD, each road user is represented
by its measurement extent across range, azimuth, and Doppler, with the
ground truth defined directly in the underlying RAD space rather than
in the projected RA or RD views discussed in
Section~\ref{sec:ra_rd_rad_representations}. Compared with center-based
RA formulations such as RODNet-style Gaussian confidence maps,
cube-level boxes preserve both spatial and Doppler extent and better
describe large, nearby, or side-viewed road users.

\subsection{World-Static Doppler Field under Ego Motion}
\label{sec:world_static_doppler_field}

\begin{figure*}[t]
\centering
\includegraphics[width=1.0\textwidth]{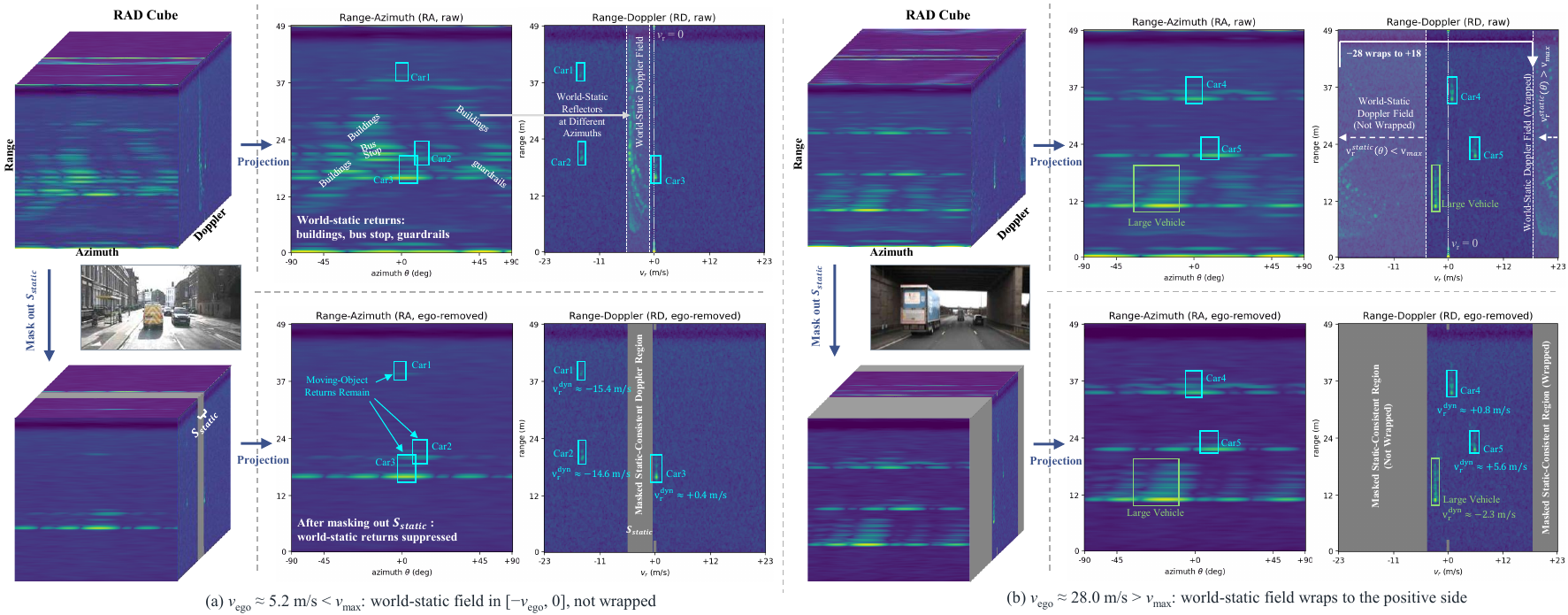}
\caption{Static-consistent Doppler-region masking at two ego speeds:
(a) $v_{\mathrm{ego}}\approx5$~m/s and
(b) $v_{\mathrm{ego}}\approx28$~m/s. The top and bottom rows show the
raw and masked RAD cubes with their corresponding RA and RD
projections, respectively. The static field forms a single Doppler
band in (a) and two disjoint wrapped bands in (b), with masked regions
shown in gray.}
\label{fig:staticmask}
\vspace{-4mm}
\end{figure*}

Building on the radial-velocity projection described above, we now
consider a moving radar observing world-static reflectors, such as
buildings, road boundaries, guardrails, and parked vehicles. These
reflectors are stationary in the world frame, so their radial velocity
relative to the radar is induced entirely by radar ego motion:
\begin{equation}
v_r^{\mathrm{static}}(\theta)
=
-
\mathbf{v}_{\mathrm{ego}}^{\top}
\mathbf{u}(\theta),
\end{equation}
where $\theta$ is the azimuth angle of the reflector relative to the
radar boresight, $\mathbf{u}(\theta)$ is the line-of-sight unit vector
from the radar to the reflector, and $\mathbf{v}_{\mathrm{ego}}$
denotes the translational velocity of the radar center expressed in
the radar body frame. Under the ideal point-sensor model, pure
rotation about the radar center changes the viewing direction but
not the distance to a fixed reflector, and therefore has no direct
radial-velocity contribution. Translation of the radar center during
turning is described by $\mathbf{v}_{\mathrm{ego}}$.

When the ego vehicle mainly moves forward, the static radial velocity
can be approximated as
\begin{equation}
v_r^{\mathrm{static}}(\theta)
\approx
-
v_{\mathrm{ego}}\cos\theta.
\end{equation}
For the forward field of view considered here,
$\cos\theta\in[0,1]$, so the world-static Doppler field lies within
$[-v_{\mathrm{ego}},0]$: world-static reflectors near the radar boresight appear
around $-v_{\mathrm{ego}}$, while their radial velocities approach zero as
$|\theta|$ approaches $90^\circ$. Since this field is mainly determined by
azimuth and is approximately independent of range, static structures
distributed across the raw RA heatmap collectively form a broad
but predictable band in the corresponding RD projection. Nearby
roadside structures often span a wider azimuth interval than
far-range regions, making this band particularly evident in
Fig.~\ref{fig:staticmask}(a), top.

Before considering Doppler wrapping, the azimuth-dependent world-static
field induces a global Doppler support over the radar field of view.
Using the Doppler-bin velocity $v_{\mathrm{D}}$, we define this support
as
\begin{equation}
\mathcal{V}_{\mathrm{static}}
=
\left\{
v_{\mathrm{D}}
\;\middle|\;
\exists\theta'\in\Theta_{\mathrm{FoV}},
\left|
v_{\mathrm{D}}
-
v_r^{\mathrm{static}}(\theta')
\right|
\leq\epsilon_v
\right\},
\end{equation}
where $\theta'$ indexes the azimuths contributing to the global
Doppler support and $\epsilon_v$ is a velocity-domain tolerance. For
cube-level masking, this support is broadcast over the complete range
and azimuth dimensions:
\begin{equation}
\mathcal{S}_{\mathrm{static}}
=
\left\{
(r,\theta,v_{\mathrm{D}})
\;\middle|\;
r\in[0,R_{\max}],
\theta\in\Theta_{\mathrm{FoV}},
v_{\mathrm{D}}\in\mathcal{V}_{\mathrm{static}}
\right\}.
\end{equation}
Here, $R_{\max}$ denotes the maximum range represented in the cube.
Thus, although $v_r^{\mathrm{static}}(\theta')$ remains
azimuth-dependent, $\mathcal{S}_{\mathrm{static}}$ contains the union
of the Doppler velocities contributed by all azimuths and applies this
global support throughout the RAD cube.

The global support is adopted because angular leakage and sidelobes of
the eight-channel virtual array spread strong static returns across
neighboring azimuths at unchanged Doppler, which a thin
azimuth-dependent region would not remove. As a result, a small
portion of moving objects, mainly road users ahead traveling in the
ego direction at a considerably lower speed, falls inside the
suppressed interval as well.

The global static-consistent region
$\mathcal{S}_{\mathrm{static}}$ defined above is generally different from the
zero-Doppler region, which only indicates a small radar-relative
radial velocity. A vehicle traveling in the same direction at a speed
close to the ego speed may therefore appear near zero Doppler despite
being independently moving, as illustrated by Car3 in
Fig.~\ref{fig:staticmask}(a), top.

The resulting global static-consistent Doppler region
$\mathcal{S}_{\mathrm{static}}$ provides a useful physical prior for
annotation, visualization, and ego-motion-aware suppression. Masking
this region suppresses most ego-motion-induced static returns, causing
roadside structures to largely disappear from the RA and RD
projections, while Car1--Car3 remain visible in the example shown in
Fig.~\ref{fig:staticmask}(a), bottom. The mask shown here is not applied
as an external preprocessing step to the RaMOD input. During inference,
RaMOD receives a sequence of full unmasked RAD cubes. Its Ego-Motion
Branch estimates frame-wise ego motion from the raw cubes and
constructs the corresponding internal Doppler suppression masks. The
resulting suppressed cube sequence is then passed to the temporal
encoder for downstream detection.

\subsection{Doppler Wrapping and Static--Dynamic Ambiguity}
\label{sec:doppler_wrapping_static_dynamic_ambiguity}

\subsubsection{Finite Doppler Interval}
\label{sec:finite_doppler_interval}

The previous subsection describes radial velocities before considering Doppler wrapping. In practice, Doppler FFT represents radial velocity only within the finite unambiguous interval $[-v_{\max},v_{\max})$. Let $v_{r,\mathrm{unw}}$ denote an unwrapped radial velocity and let $v_r$ denote the radial velocity represented in the radar cube.

When $v_{r,\mathrm{unw}}\in[-v_{\max},v_{\max})$, no wrapping occurs and
\begin{equation}
v_r
=
v_{r,\mathrm{unw}}.
\end{equation}
Otherwise, Doppler wrapping maps the velocity back into this interval:
\begin{equation}
\label{eq:doppler_wrap}
v_r
=
\mathrm{wrap}(v_{r,\mathrm{unw}};v_{\max})
=
\left(
(v_{r,\mathrm{unw}}+v_{\max}) \bmod 2v_{\max}
\right)
-
v_{\max}.
\end{equation}

\subsubsection{Moving-Object Wrapping}
\label{sec:dynamic_object_wrapping}

\begin{figure}[t]
\centering
\includegraphics[width=\columnwidth]{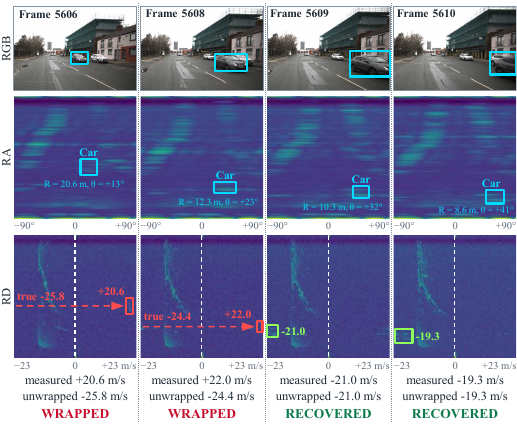}
\caption{Moving-object Doppler wrapping over four consecutive frames
($v_{\mathrm{ego}}\approx14$~m/s; RGB, RA, and RD from top to bottom).
The oncoming vehicle is wrapped to the positive Doppler side in
the first two frames, but returns to the negative side as its
increasing azimuth reduces the radial-speed magnitude below
$v_{\max}$.}
\label{fig:dynwrap}
\vspace{-6mm}
\end{figure}

For an independently moving object, the radial velocity represented
in the radar cube is
\begin{equation}
v_r^{\mathrm{dyn}}(\theta)
=
\mathrm{wrap}
\left(
v_{r,\mathrm{unw}}^{\mathrm{dyn}}(\theta);
v_{\max}
\right).
\end{equation}
Here, $v_{r,\mathrm{unw}}^{\mathrm{dyn}}(\theta)$ denotes the
object's unwrapped radial velocity relative to the radar.

Whether wrapping occurs is strongly related to traffic direction.
For an object near the radar boresight, i.e.,
$\theta_o\approx0$, let $v_{r,\mathrm{world}}^{\mathrm{dyn}}$
denote the radial component of its ground-referenced velocity along
the radar line of sight. The magnitude of its unwrapped
radar-relative radial velocity can be approximated as
\begin{equation}
\left|
v_{r,\mathrm{unw}}^{\mathrm{dyn}}
\right|
\approx
\begin{cases}
v_{\mathrm{ego}}
+
\left|
v_{r,\mathrm{world}}^{\mathrm{dyn}}
\right|,
& \text{oncoming vehicle},\\[4pt]
\left|
v_{\mathrm{ego}}
-
v_{r,\mathrm{world}}^{\mathrm{dyn}}
\right|,
& \text{same-direction vehicle}.
\end{cases}
\end{equation}

Oncoming vehicles are therefore more likely to exceed $v_{\max}$,
whereas same-direction vehicles are less likely to wrap unless their
relative radial speed is large. This behavior is illustrated in
Fig.~\ref{fig:dynwrap}. In the first two frames, the oncoming vehicle
has unwrapped radial velocities of approximately
$-25.8$ and $-24.4~\mathrm{m/s}$, which exceed the unambiguous range
and are measured at approximately $+20.6$ and $+22.0~\mathrm{m/s}$ on
the opposite side of the Doppler axis. As the vehicle azimuth
increases, its radial projection decreases; in the last two frames,
the radial velocities of approximately $-21.0$ and
$-19.3~\mathrm{m/s}$ fall within the unambiguous interval and are
measured without wrapping. Moving-object wrapping therefore depends
on traffic direction, ego speed, object speed, azimuth, and radar
Doppler coverage.

\subsubsection{Ego-Motion-Induced Static-Field Wrapping}
\label{sec:static_field_wrapping}

The same wrapping effect can occur for the ego-motion-induced
world-static Doppler field. When the ego speed is large relative to
the radar Doppler coverage, the unwrapped static radial velocity
$v_r^{\mathrm{static}}(\theta)$ may exceed the measurable Doppler
interval. The corresponding wrapped static field represented in the radar
cube becomes
\begin{equation}
v_{r,\mathrm{wrap}}^{\mathrm{static}}(\theta)
=
\mathrm{wrap}
\left(
v_r^{\mathrm{static}}(\theta);
v_{\max}
\right).
\end{equation}

This wrapping does not clip the static field at the velocity boundary.
Instead, velocities outside the measurable interval are folded to the
opposite side of the Doppler axis. Consequently, at high ego speeds,
the originally continuous static field may appear as a discontinuous
pattern in the RD projection or RAD cube, with static returns split
across the negative and positive Doppler regions. In
Fig.~\ref{fig:staticmask}(b), for example, the boresight static
component at approximately $-28~\mathrm{m/s}$ wraps to
$+18~\mathrm{m/s}$.

For masking in the measured RAD cube, the global Doppler support
$\mathcal{V}_{\mathrm{static}}$ must therefore be constructed from
$v_{r,\mathrm{wrap}}^{\mathrm{static}}(\theta)$ rather than
$v_r^{\mathrm{static}}(\theta)$. Accordingly,
$\mathcal{V}_{\mathrm{static}}$, and hence
$\mathcal{S}_{\mathrm{static}}$, may separate into non-wrapped and
wrapped parts, producing two disjoint Doppler bands for masking, as
illustrated in the lower row of Fig.~\ref{fig:staticmask}(b).

\subsubsection{Static--Dynamic Overlap after Wrapping}
\label{sec:static_dynamic_overlap_after_wrapping}

In a moving-ego scenario, world-static reflectors form the
azimuth-dependent ego-motion-induced static Doppler field
$v_r^{\mathrm{static}}(\theta)$. Moving objects, particularly
oncoming vehicles, may have large radar-relative radial velocities
and are therefore more likely to exceed the unambiguous Doppler
range and undergo Doppler wrapping.

A wrapping-induced overlap occurs when
\begin{equation}
\min_{k\in\mathbb{Z}\setminus\{0\}}
\left|
v_{r,\mathrm{unw}}^{\mathrm{dyn}}
-
v_r^{\mathrm{static}}(\theta_o)
-
2k v_{\max}
\right|
\leq
\epsilon_v,
\end{equation}
where $\theta_o$ is the representative object azimuth. This condition
indicates that the radial-velocity difference between the moving
object and the static Doppler field is close to a non-zero integer
multiple of the Doppler wrapping period $2v_{\max}$.

When $v_{\max}$ is small, the wrapping period is also small, making
common traffic velocities more likely to be folded into or close to
the static-consistent Doppler region. In this case, Doppler alone no
longer provides sufficient evidence to distinguish the moving object
from world-static returns.

The range and azimuth dimensions of the full RAD cube may still
provide additional spatial information. However, finite angular
resolution, beamforming sidelobes, and angular leakage can spread
strong static returns into neighboring azimuth regions while
preserving their Doppler values. Consequently, static--dynamic
ambiguity may also reduce separability in the RAD cube.

This limitation motivates the radar-configuration and Doppler-coverage
discussion in the following subsection.

\subsection{Radar Configuration and Doppler Coverage Design}
\label{sec:doppler_coverage_design}

The above analysis shows that radar configuration directly affects the separability between
dynamic-object returns and the ego-motion-induced static background.
First, to avoid static-field wrapping, the radar configuration should
satisfy
\begin{equation}
v_{\max}
>
\max_{\theta\in\Theta_{\mathrm{FoV}}}
\left|
v_r^{\mathrm{static}}(\theta)
\right|.
\end{equation}
For approximately forward ego motion over the intended operating
range, a sufficient condition is
\begin{equation}
v_{\max}
>
v_{\mathrm{ego}}^{\max},
\end{equation}
where $v_{\mathrm{ego}}^{\max}$ denotes the maximum ego speed
considered.

Second, a local condition for separating dynamic and static returns
at the same azimuth is
\begin{equation}
\min_{k\in\mathbb{Z}}
\left|
v_{r,\mathrm{world}}^{\mathrm{dyn}}
-
2k v_{\max}
\right|
>
\delta_v,
\end{equation}
where $\delta_v$ accounts for Doppler resolution, object-level Doppler
spread, angular leakage, and the tolerance used to define
$\mathcal{S}_{\mathrm{static}}$. This local condition alone does not
ensure separation from the global suppression region, which also
includes static Doppler contributions from other azimuths.
The required Doppler coverage therefore depends on the target
scenario, including oncoming traffic, high ego speed, and overtaking
vehicles.

In practice, $v_{\max}$ is constrained by the radar configuration and is determined by the effective slow-time sampling interval:
\begin{equation}
v_{\max}
=
\frac{\lambda}{4T_{\mathrm{eff}}},
\end{equation}
where $\lambda$ is the radar wavelength and $T_c$ denotes the chirp
repetition interval. For a single-transmitter radar,
$T_{\mathrm{eff}}=T_c$. For a standard TDM-MIMO radar with
$N_{\mathrm{TX}}$ transmitters,
\begin{equation}
T_{\mathrm{eff}}
=
N_{\mathrm{TX}}T_c,
\qquad
v_{\max}
=
\frac{\lambda}{4N_{\mathrm{TX}}T_c}.
\end{equation}

A shorter chirp repetition interval increases $v_{\max}$, whereas using
more transmit antennas in standard TDM-MIMO reduces it. Radar
configuration must therefore balance angular resolution, Doppler
resolution, frame rate, and unambiguous Doppler coverage to preserve
static--dynamic separability.

\subsection{Object-Level Velocity Recovery}
\label{sec:object_level_velocity_recovery}

\begin{figure}[t]
\centering
\includegraphics[width=0.85\columnwidth]{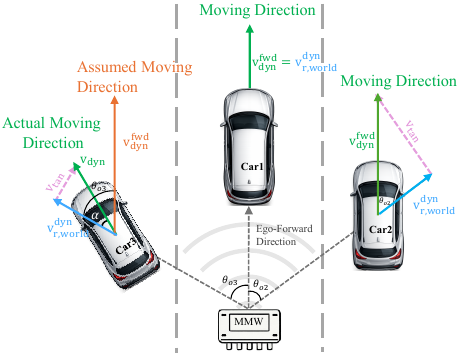}
\caption{Object-level forward-velocity recovery from a single-radar
radial measurement. Car~1 lies near the radar boresight, while the
off-axis Car~2 requires azimuth-based correction. For Car~3,
the actual projection angle $\alpha$ differs from the assumed angle
$\theta_{o3}$, resulting in a biased velocity estimate. The
tangential component $v_{\mathrm{tan}}$ is not directly
observable from Doppler.}
\label{fig:velrecovery}
\vspace{-3mm}
\end{figure}

Radar Doppler directly measures radial velocity rather than full
Cartesian velocity. For each object-level box in the radar cube, an
object-level radial velocity $v_r^{\mathrm{dyn}}$ is estimated from
the Doppler distribution within the box rather than from a single
Doppler peak, since one object may occupy multiple Doppler bins. When
Doppler wrapping occurs,
$v_{r,\mathrm{unw}}^{\mathrm{dyn}}$ denotes the recovered unwrapped
radial velocity, following the notation introduced in
Section~\ref{sec:dynamic_object_wrapping}. After resolving any Doppler wrapping, ego motion is compensated to
obtain the object's ground-referenced radial component:

\begin{equation}
v_{r,\mathrm{world}}^{\mathrm{dyn}}
=
v_{r,\mathrm{unw}}^{\mathrm{dyn}}
+
\mathbf{v}_{\mathrm{ego}}^{\top}
\mathbf{u}(\theta_o),
\end{equation}
where $\theta_o$ is the representative object azimuth and
$\mathbf{u}(\theta_o)$ is the line-of-sight unit vector. The quantity
$v_{r,\mathrm{world}}^{\mathrm{dyn}}$ remains a radial component,
while the orthogonal tangential component
$v_{\mathrm{tan}}$ is not observable from the radar Doppler measurements, as illustrated in Fig.~\ref{fig:velrecovery}.

To estimate the forward velocity, nearby vehicles are assumed to move
approximately along the ego-forward direction, as illustrated by
Car~1 and Car~2 in Fig.~\ref{fig:velrecovery}. Under this road-scene
assumption, with approximately forward ego motion and a forward-facing radar,
\begin{equation}
v_{r,\mathrm{world}}^{\mathrm{dyn}}
\approx
v_{\mathrm{dyn}}^{\mathrm{fwd}}\cos\theta_o,
\end{equation}
giving
\begin{equation}
v_{\mathrm{dyn}}^{\mathrm{fwd}}
\approx
\frac{
v_{r,\mathrm{world}}^{\mathrm{dyn}}
}{
\cos\theta_o
}
=
\frac{
v_{r,\mathrm{unw}}^{\mathrm{dyn}}
+
v_{\mathrm{ego}}\cos\theta_o
}{
\cos\theta_o
}.
\end{equation}
Here, $v_{\mathrm{dyn}}^{\mathrm{fwd}}$ denotes the
ground-referenced object velocity along the ego-forward direction.
For Car~1, $\theta_o\approx0$ and the forward and radial components
are approximately equal, whereas the off-axis Car~2 requires the
azimuth correction. When wrapping does not occur,
$v_r^{\mathrm{dyn}}$ can be used directly in place of
$v_{r,\mathrm{unw}}^{\mathrm{dyn}}$.

This recovery becomes ill-conditioned as $\theta_o$ approaches
$90^\circ$. It also becomes biased when the actual motion deviates
from the assumed ego-forward direction. For Car~3 in
Fig.~\ref{fig:velrecovery}, the actual velocity forms an angle
$\alpha$ with the line of sight, whereas the recovery assumes the
angle $\theta_{o3}$. When $\alpha\neq\theta_{o3}$, the radial
measurement alone is insufficient to recover the actual object
velocity accurately, as may occur during crossing or turning.

The preceding analysis shows that ego motion induces a predictable
static Doppler field, providing a physical prior for background
suppression. Spatial--Doppler structure and temporal changes provide
complementary cues for moving-object detection, while Doppler
wrapping requires explicit treatment in velocity estimation.
These observations motivate RaMOD's ego-motion estimation and
internal Doppler suppression, temporal cube encoding, and RAD-space
detection with wrapping-aware velocity estimation, presented in
the following section.

\begin{figure*}[!t]
\centering
\includegraphics[width=1.0\textwidth]{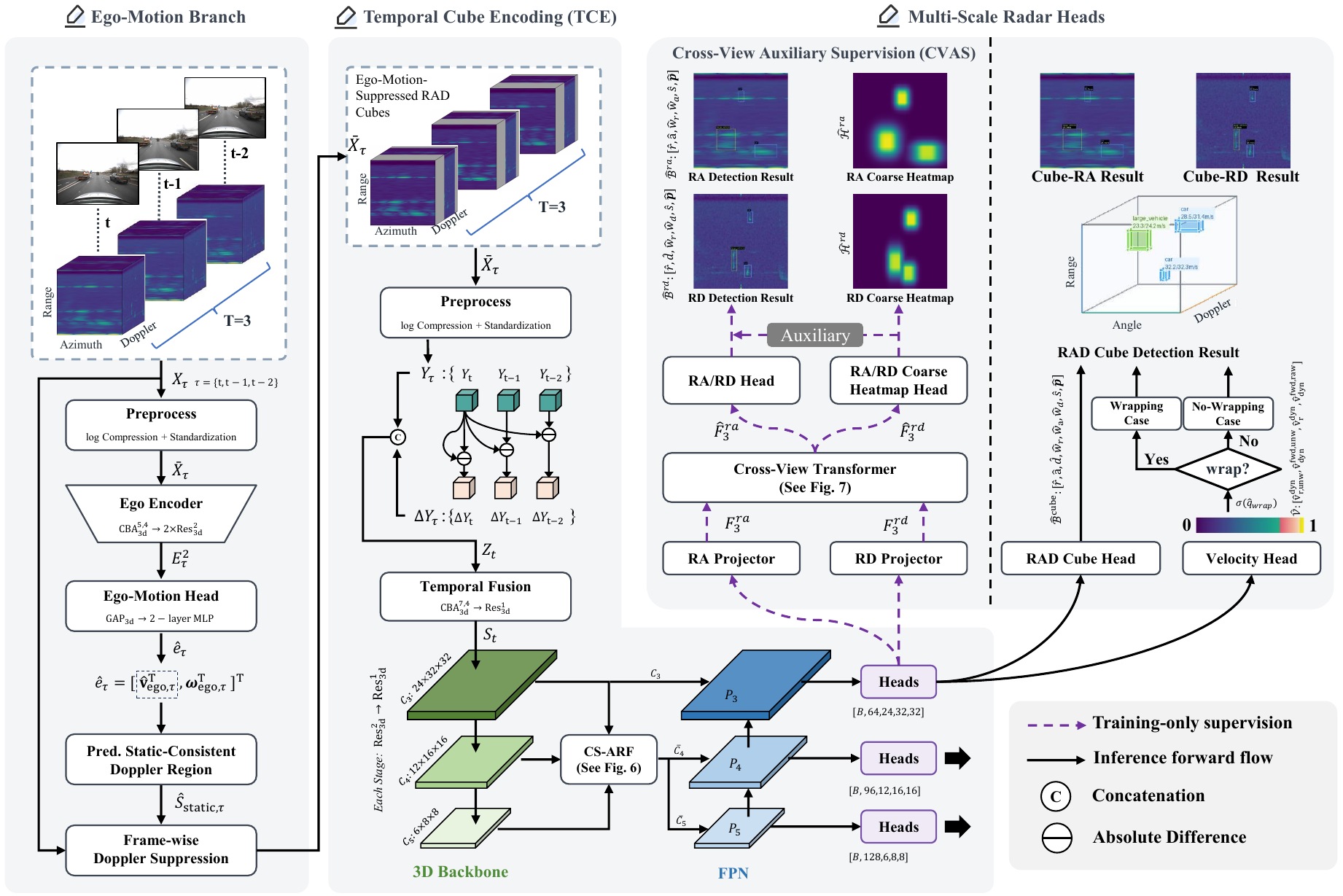}
\caption{Overview of RaMOD. The forward path consists of an Ego-Motion Branch, a Temporal Cube Encoding Branch, and Multi-Scale Radar Heads. The Ego-Motion Branch estimates frame-wise ego motion and suppresses static-consistent Doppler responses. The temporal branch encodes the resulting cube sequence into a multi-scale pyramid, from which the radar heads predict cube boxes, object-level radial and forward velocities, and CVAS outputs including RA/RD boxes and coarse object-support heatmaps.}
\label{fig:method_overview}
\vspace{-3mm}
\end{figure*}

\section{Method}
\label{sec:method}

\subsection{Overview}
\label{sec:method_overview}

RaMOD performs moving-object detection and velocity estimation from
raw range--azimuth--Doppler (RAD) cubes acquired by a moving mmWave
radar. As shown in Fig.~\ref{fig:method_overview}, it consists of
(a) an Ego-Motion Branch, (b) a Temporal Cube Encoding Branch, and
(c) Multi-Scale Radar Heads.

Let $\mathcal{X}_{t}$ denote a temporal window of $T$ raw RAD cubes,
where $t$ is the target-frame time and $\tau$ indexes frames within
the window.
Each cube $X_{\tau}\in\mathbb{R}^{R\times A\times D}$ has range,
azimuth, and Doppler dimensions $R$, $A$, and $D$, respectively.
Unless otherwise stated, $T=3$.

\textbf{(a) Ego-Motion Branch.}
Each unmasked cube is processed independently to estimate
$\hat{\mathbf{e}}_{\tau}=\operatorname{EgoMotion}(X_{\tau})$.
Following the static-Doppler model in
Sec.~\ref{sec:world_static_doppler_field}, the predicted radar
translational velocity defines the static-consistent Doppler region
$\mathcal{S}_{\mathrm{static}}$ for internal suppression,
giving $\tilde{X}_{\tau}=
\operatorname{Suppress}(X_{\tau},\hat{\mathbf{e}}_{\tau})$.
Applied frame-wise, this reduces ego-motion-induced background
responses before detection and produces the suppressed sequence
$\tilde{\mathcal{X}}_{t}$ for temporal encoding. The suppression
operation is detailed in Sec.~\ref{sec:ego_motion_compensation}.

\textbf{(b) Temporal Cube Encoding Branch.}
The Temporal Cube Encoding Branch converts the suppressed sequence
into a motion-aware temporal representation and encodes it with a
3D backbone, CS-ARF, and a cube feature pyramid, producing
$\mathcal{P}_{t}=\operatorname{TCE}(\tilde{\mathcal{X}}_t)$.
These multi-scale features comprise $P^{(t)}_5$, $P^{(t)}_4$, and
$P^{(t)}_3$ for the target frame at time $t$. For readability, the
time superscript of each pyramid level is omitted when the context
is clear.

\textbf{(c) Multi-Scale Radar Heads.}
Multi-Scale Radar Heads operate on all pyramid levels. At each level,
a Cube Detection Head predicts RAD cube boxes, objectness, and class
probabilities; a Velocity Head predicts raw-Doppler and
Doppler-unwrapped radial/forward velocity pairs and a wrapping
probability for each cube-box candidate; and Cross-View Auxiliary
Supervision (CVAS) branches predict RA/RD boxes and coarse
object-support heatmaps.
The complete head output is
$\hat{\mathcal{O}}_{t}=\operatorname{Head}(\mathcal{P}_{t})$,
where $\hat{\mathcal{B}}^{cube}$ denotes RAD cube detections,
$\hat{\mathcal{B}}^{ra}$ and $\hat{\mathcal{B}}^{rd}$ denote CVAS box
predictions, $\hat{\mathcal{H}}^{ra}$ and
$\hat{\mathcal{H}}^{rd}$ denote CVAS coarse object-support heatmaps,
and $\hat{\mathcal{V}}$ denotes object-level velocity predictions.
At inference, coarse-support gating, proposal decoding, NMS, and
wrapping-based velocity selection convert these outputs into final
moving-object states.

The three components are first optimized jointly using the multi-task
objective. The Velocity Heads are subsequently fine-tuned with all
other network components frozen. The remaining subsections detail
the three components, the inference procedure, and the training
objective.

\subsection{Ego-Motion Branch}
\label{sec:ego_motion_compensation}

Following the analysis in
Sec.~\ref{sec:world_static_doppler_field}, the Ego-Motion Branch
estimates frame-wise ego motion independently from each full unmasked
RAD cube. The predicted ego translational velocity is then used to construct an
internal cube-level static-consistent Doppler suppression mask. Applied
frame-wise, this operation produces the suppressed cube sequence
$\tilde{\mathcal{X}}_t$ for subsequent temporal encoding.

For clarity, we define the convolutional notation used in the following
sections. A 3D convolutional unit is
\begin{equation}
\mathrm{CBA}_{3d}^{k,s}(\mathbf{x})
=
\mathrm{SiLU}\left(
\mathrm{BN}\left(
\mathrm{Conv3D}_{k,s}(\mathbf{x})
\right)
\right),
\end{equation}
where $k$ and $s$ denote kernel size and stride and
$\mathrm{SiLU}(z)=z/(1+e^{-z})$ is the sigmoid linear unit
activation.
A residual 3D block is
\begin{equation}
\mathrm{Res}_{3d}^{s}(\mathbf{x})
=
\mathrm{SiLU}\!\left(
\mathcal{F}_{3d}^{s}(\mathbf{x})
+
\operatorname{Skip}^{s}(\mathbf{x})
\right),
\end{equation}
where $\mathcal{F}_{3d}^{s}$ consists of
$\mathrm{CBA}_{3d}^{3,s}$ followed by a batch-normalized
$3\times3\times3$ convolution, and
$\operatorname{Skip}^{s}$ is either an identity
shortcut or a $1\times1\times1$ convolution when spatial resolution
or channel dimension changes.

The Ego-Motion Branch applies logarithmic compression to each raw
RAD cube to reduce its dynamic range and limit the numerical dominance
of strong returns. Each cube is then standardized independently and
processed by a shared lightweight 3D encoder and a global pooling head. For frame $\tau$, the preprocessed cube is $\bar{X}_{\tau}\in\mathbb{R}^{1\times R\times A\times D}$. The ego encoder is implemented as
\begin{equation}
\begin{aligned}
E_{\tau}^{0} &= \mathrm{CBA}_{3d}^{5,4}(\bar{X}_{\tau}),\\
E_{\tau}^{1} &= \mathrm{Res}_{3d}^{2}(E_{\tau}^{0}),\\
E_{\tau}^{2} &= \mathrm{Res}_{3d}^{2}(E_{\tau}^{1}).
\end{aligned}
\end{equation}

Applying 3D global average pooling gives
$\mathbf{g}_{\tau}=\operatorname{GAP}_{3d}(E_{\tau}^{2})$,
which a two-layer MLP maps to the six-dimensional ego-motion vector:
\begin{equation}
\hat{\mathbf{e}}_{\tau}
=
W_2\,\mathrm{Dropout}
\left(
\mathrm{SiLU}(W_1\mathbf{g}_{\tau})
\right).
\end{equation}

The output contains both translational and angular components of ego
motion expressed in the radar body frame,
\begin{equation}
\hat{\mathbf{e}}_{\tau}
=
\left[
\hat{\mathbf{v}}_{\mathrm{ego},\tau}^{\top},
\hat{\boldsymbol{\omega}}_{\mathrm{ego},\tau}^{\top}
\right]^{\top}.
\end{equation}
Both components are jointly supervised as ego-motion outputs, whereas
only the translational velocity
$\hat{\mathbf{v}}_{\mathrm{ego},\tau}$ is used to construct the
static-consistent Doppler suppression mask.

For each frame, the predicted ego translational velocity
$\hat{\mathbf{v}}_{\mathrm{ego},\tau}$, expressed in the radar body
frame, is substituted into the world-static Doppler model in
Sec.~\ref{sec:world_static_doppler_field} to obtain the predicted
static radial-velocity field
$\hat{v}_{r,\tau}^{\mathrm{static}}(\theta)$.

Following the Doppler-wrapping and global-support construction in
Secs.~\ref{sec:world_static_doppler_field} and
\ref{sec:static_field_wrapping}, the predicted static field
$\hat{v}_{r,\tau}^{\mathrm{static}}(\theta)$ is converted into the
frame-wise cube-level suppression region
$\hat{\mathcal{S}}_{\mathrm{static},\tau}$. In implementation, the
velocity tolerance $\epsilon_v$ is realized by padding $p$ Doppler
bins. The suppressed cube is
\begin{equation}
\tilde{X}_{\tau}(r,\theta,v_{\mathrm{D}})
=
\begin{cases}
\epsilon_x,
&
(r,\theta,v_{\mathrm{D}})
\in
\hat{\mathcal{S}}_{\mathrm{static},\tau},
\\
X_{\tau}(r,\theta,v_{\mathrm{D}}),
&
\text{otherwise},
\end{cases}
\end{equation}
where $\epsilon_x$ is a small floor value.

\subsection{Temporal Cube Encoding Branch}
\label{sec:temporal_cube_encoding}

\begin{figure}[!t]
\centering
\includegraphics[width=0.9\linewidth]{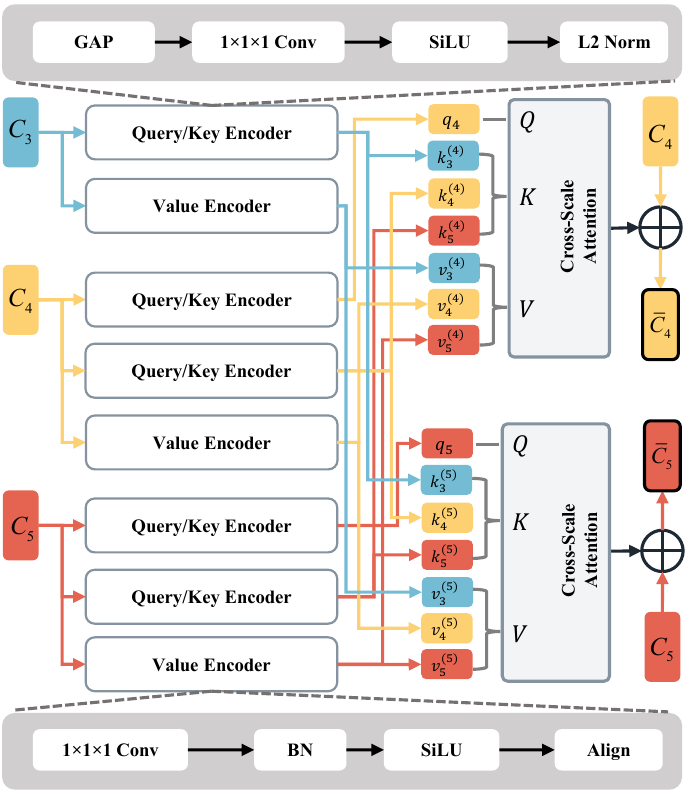}

\caption{Cross-Scale Attentive Residual Fusion (CS-ARF).
For each independent target path ($C_i\in\{C_4,C_5\}$),
$C_i$ supplies a query; $\{C_3,C_4,C_5\}$ supply keys and
spatially aligned values. Attention-weighted aggregation,
3D refinement, and residual scaling produce an update
added to $C_i$ to obtain $\bar{C}_i$.}

\label{fig:cs_arf}
\vspace{-3mm}
\end{figure}

After ego-motion-aware Doppler suppression, each cube is log-compressed and sample-wise standardized as
$Y_{\tau}=\mathrm{Norm}(\log_{10}(\tilde{X}_{\tau}+1))$. A single cube describes the instantaneous radar response, but independently moving objects are more reliably distinguished by changes in their RAD support over time. RaMOD therefore represents the temporal window with both the standardized cubes and their differences from the current frame. For each frame in the window, we compute
\begin{equation}
\Delta Y_{\tau}=|Y_{\tau}-Y_t|,
\end{equation}
and concatenate $\{Y_{\tau}\}$ and $\{\Delta Y_{\tau}\}$ along the channel dimension to form $Z_t$. The standardized cubes preserve the absolute range--azimuth--Doppler evidence, whereas the difference cubes emphasize temporally varying responses. This complementary representation helps the encoder separate persistent static structure from object-related motion cues.

The fused temporal cube is first processed by a 3D stem:
\begin{equation}
S_t
=
\mathrm{Res}_{3d}^{1}
\left(
\mathrm{CBA}_{3d}^{7,4}(Z_t)
\right).
\end{equation}
Because $Z_t$ contains $T$ ego-motion-suppressed cubes and $T$ difference cubes, the stem input has $2T$ channels. The stride-$4$ 3D convolution reduces the dense RAD tensor early while retaining local neighborhoods in range, azimuth, and Doppler.

The backbone contains three residual 3D stages:
\begin{equation}
\begin{aligned}
C_3 &= \mathrm{Res}_{3d}^{1}
      \left(\mathrm{Res}_{3d}^{2}(S_t)\right),\\
C_4 &= \mathrm{Res}_{3d}^{1}
      \left(\mathrm{Res}_{3d}^{2}(C_3)\right),\\
C_5 &= \mathrm{Res}_{3d}^{1}
      \left(\mathrm{Res}_{3d}^{2}(C_4)\right).
\end{aligned}
\end{equation}

The three stages progressively enlarge the RAD-space receptive field
and reduce spatial resolution, combining fine local evidence in
$C_3$ with broader context in $C_4$ and $C_5$. This supports distant
pedestrians occupying few bins and nearby side-viewed vehicles
spanning wider azimuth and Doppler regions.

\textbf{Cross-Scale Attentive Residual Fusion.}
To exchange these complementary features before pyramid decoding,
RaMOD introduces Cross-Scale Attentive Residual Fusion (CS-ARF).
As shown in Fig.~\ref{fig:cs_arf}, CS-ARF instantiates two independent
target-stage paths for $C_4$ and $C_5$. The high-resolution feature
$C_3$ is retained unchanged and later serves as the lateral feature
of the finest pyramid level.

For each target $i\in\{4,5\}$, CS-ARF reads all source stages $j\in\{3,4,5\}$. The target feature produces a query, and every source stage produces a key. Both encoders perform 3D global average pooling, a $1\times1\times1$ channel projection, a SiLU activation, and $L_2$ normalization:
\begin{equation}
\mathbf{q}_{i}=\operatorname{Norm}(\psi_i^{Q}(C_i)),
\quad
\mathbf{k}_{ij}=\operatorname{Norm}(\psi_{ij}^{K}(C_j)).
\end{equation}
The normalized query--key similarity gives the cross-scale attention weight from source stage $j$ to target stage $i$:
\begin{equation}
\alpha_{ij}
=
\operatorname{softmax}_{j}
\left(
\frac{\mathbf{q}_{i}^{\top}\mathbf{k}_{ij}}{\sqrt{d_h}}
\right).
\end{equation}
where $d_h$ is the query/key embedding dimension.

Each source also provides a value feature through a target-specific $1\times1\times1$ projection. The value is adaptively pooled or trilinearly upsampled to the target resolution before attention-weighted aggregation. The target update is
\begin{equation}
\bar{C}_{i}
=
C_i
+
\tanh(\gamma_i)\,
\rho_i
\left(
\sum_j \alpha_{ij}\,
\operatorname{Align}_{i}(\phi_{ij}(C_j))
\right),
\end{equation}
where $\psi_i^{Q}$ and $\psi_{ij}^{K}$ denote the target-specific query and key encoders, respectively; $\phi_{ij}$ is the target-specific $1\times1\times1$ value projection; $\rho_i$ is the subsequent 3D convolutional refinement block; and $\gamma_i$ is a learnable residual-scale parameter. The bounded $\tanh(\gamma_i)$ factor initially limits the cross-scale update, while the identity path preserves the target-stage representation.

\textbf{Cube feature pyramid.}
A multi-scale 3D feature pyramid is then constructed from the CS-ARF outputs. Let $\bar{C}_3=C_3$, while $\bar{C}_4$ and $\bar{C}_5$ are the two refined target stages. The top-down pyramid is
\begin{equation}
\begin{aligned}
P_5 &= \mathrm{CBA}_{3d}^{3,1}
      \left(\mathrm{CBA}_{3d}^{1,1}(\bar{C}_5)\right),\\
P_4 &= \eta_4
      \left(
      [\operatorname{Up}(P_5),\mathrm{CBA}_{3d}^{1,1}(\bar{C}_4)]
      \right),\\
P_3 &= \eta_3
      \left(
      [\operatorname{Up}(P_4),\mathrm{CBA}_{3d}^{1,1}(\bar{C}_3)]
      \right),
\end{aligned}
\end{equation}
where $\operatorname{Up}(\cdot)$ is trilinear interpolation, $[\cdot,\cdot]$ denotes channel concatenation, and $\eta_3,\eta_4$ are two-layer $3\times3\times3$ convolutional refinement blocks. Thus, $P_5$ carries the deepest semantic context, $P_4$ combines it with the refined middle-scale representation, and $P_3$ restores fine-scale evidence through $C_3$. The resulting pyramid supports multi-scale cube localization, cross-view projection, and velocity estimation.

\subsection{Multi-Scale Radar Heads}
\label{sec:multi_scale_heads}

RaMOD attaches task-specific heads to all pyramid features $\mathcal{P}_t=\{P_5,P_4,P_3\}$, rather than only using the deepest feature map. This section describes the multi-scale RAD cube detection and velocity heads, followed by the cross-view auxiliary branches for RA/RD detection and coarse object support.

\subsubsection{Anchor parameterization and assignment}
RaMOD uses radar-aware anchors in the RAD cube space. Let $\mathcal{A}=\{\mathbf{a}_m\}_{m=1}^{N_{\mathrm{A}}}$ denote the set of 3D anchor sizes, where $\mathbf{a}_m=(a^r_m,a^a_m,a^d_m)$ specifies the range, azimuth, and Doppler extent. The six anchors $(9,12,8)$, $(15,19,12)$, $(15,28,17)$, $(17,30,32)$, $(18,55,23)$, and $(20,57,47)$ are assigned two per pyramid level, from the largest at $P_5$ to the smallest at $P_3$.

For the auxiliary views, the 3D anchors are projected into view-specific anchors:
\begin{equation}
\begin{aligned}
\mathcal{A}^{ra}_{\ell}
&=
\{(a^r,a^a)\mid(a^r,a^a,a^d)\in\mathcal{A}_{\ell}\},
\\
\mathcal{A}^{rd}_{\ell}
&=
\{(a^r,a^d)\mid(a^r,a^a,a^d)\in\mathcal{A}_{\ell}\}.
\end{aligned}
\end{equation}
Thus, cube, RA, and RD heads share a consistent anchor origin while predicting boxes in their corresponding measurement spaces.

For training, each ground-truth cube box is assigned to the grid cell containing its center at every pyramid level. Its size is compared with the anchors of that level using 3D IoU in the local grid coordinate system. Anchors whose IoU is larger than the matching threshold are marked as positive. If no anchor exceeds the threshold, the anchor with the maximum IoU is selected to ensure that each object has at least one positive assignment. The RA and RD labels follow the same rule using the projected 2D boxes and 2D anchors. Velocity targets are attached to positive cube anchors only.

\subsubsection{RAD cube detection head}

The anchor-based 3D detection head at each pyramid level
$\ell\in\{5,4,3\}$ predicts
\begin{equation}
\hat{\mathbf{b}}^{\mathrm{cube}}_{\ell}
=
\operatorname{CubeHead}_{\ell}(P_{\ell}).
\end{equation}

After a $1\times1\times1$ stem, separate regression and classification
towers predict the RAD box center, extent, and objectness $\hat{s}$,
and class probabilities $\hat{\mathbf{p}}$, respectively. Doppler
remains an explicit detection dimension. Precise cube-box regression
is challenging for sparse, extended returns mixed with sidelobes or
ego-motion-induced static responses, especially for weak road users
or partially visible RA/RD support.

For a grid cell $\mathbf{g}=(g_r,g_a,g_d)$ and an anchor
$\mathbf{a}_k=(a^r_k,a^a_k,a^d_k)$ at level $\ell$, the decoded cube box is
\begin{equation}
\hat{\mathbf{c}}
=
(\sigma(\mathbf{o})+\mathbf{g})\odot\mathbf{s}_{\ell},
\quad
\hat{\mathbf{q}}
=
\exp(\mathbf{u})\odot\mathbf{a}_k,
\end{equation}
where $\mathbf{o}$ and $\mathbf{u}$ are the predicted center and size logits,
$\mathbf{s}_{\ell}$ is the stride from the feature grid to the input RAD cube,
$\hat{\mathbf{c}}=(\hat{r},\hat{a},\hat{d})$ is the box center, and
$\hat{\mathbf{q}}=(\hat{w}_r,\hat{w}_a,\hat{w}_d)$ is the box extent. Objectness
and class probabilities are obtained by sigmoid activations.

\subsubsection{Object-level Velocity Head}
Beyond object localization, RaMOD estimates object-level Doppler-related velocity attributes from cube features. For each cube anchor, the Velocity Head predicts two velocity pairs and a Doppler-wrapping probability:

\begin{equation}
\hat{\mathbf{y}}^{\mathrm{vel}}_{\ell}
=
\operatorname{VelocityHead}_{\ell}(P_{\ell}).
\end{equation}

The output contains an unwrapped radial-velocity/forward-velocity pair
$\bigl(\hat{v}_{r,\mathrm{unw}}^{\mathrm{dyn}},
\hat{v}_{\mathrm{dyn}}^{\mathrm{fwd,unw}}\bigr)$ and a raw-Doppler
radial-velocity/forward-velocity pair
$\bigl(\hat{v}_{r}^{\mathrm{dyn}},
\hat{v}_{\mathrm{dyn}}^{\mathrm{fwd,raw}}\bigr)$. The forward-velocity
components follow the definition in
Sec.~\ref{sec:object_level_velocity_recovery}. The head also predicts
a wrapping logit $\hat{q}_{\mathrm{wrap}}$, with
$\hat{p}_{\mathrm{wrap}}=\sigma(\hat{q}_{\mathrm{wrap}})$.

During inference, RaMOD selects the unwrapped estimate when the predicted
wrapping probability is high and otherwise uses the raw-Doppler
estimate:
\begin{equation}
\hat{\mathbf{v}}
=
\begin{cases}
\hat{\mathbf{v}}^{\mathrm{unw}},
& \hat{p}_{\mathrm{wrap}}\geq0.5,\\
\hat{\mathbf{v}}^{\mathrm{raw}},
& \hat{p}_{\mathrm{wrap}}<0.5.
\end{cases}
\end{equation}

The Velocity Head is attached to the same multi-scale cube features
$\{P_5,P_4,P_3\}$ as the cube detector.
This allows the model to preserve the reliability of
raw Doppler measurements in normal cases while recovering unwrapped
velocity estimates when Doppler wrapping occurs.

\subsubsection{Cross-View Auxiliary Supervision}
\label{sec:cross_view_auxiliary}

To provide complementary geometric evidence for this ambiguous 3D
regression, RaMOD introduces Cross-View Auxiliary Supervision (CVAS),
which exchanges spatial and Doppler cues between RA and RD views and
provides sparse RA/RD box supervision and dense object-support
heatmaps. The 3D cube features are first projected into these views
using mean and max aggregation along the corresponding dimension:

\begin{equation}
\begin{aligned}
F^{ra}_{\ell}
&=
\rho^{ra}_{\ell}
\left(
\left[
\operatorname{Mean}_{D}(P_{\ell}),
\operatorname{Max}_{D}(P_{\ell})
\right]
\right),\\
F^{rd}_{\ell}
&=
\rho^{rd}_{\ell}
\left(
\left[
\operatorname{Mean}_{A}(P_{\ell}),
\operatorname{Max}_{A}(P_{\ell})
\right]
\right),
\end{aligned}
\end{equation}

where $\rho^{ra}_{\ell}$ and $\rho^{rd}_{\ell}$ consist of a $1\times1$ convolution followed by depthwise-separable $3\times3$ refinement layers. The RA view preserves spatial layout, while the RD view preserves range--Doppler motion structure.

\begin{figure}[!t]
\centering
\includegraphics[width=0.85\linewidth]{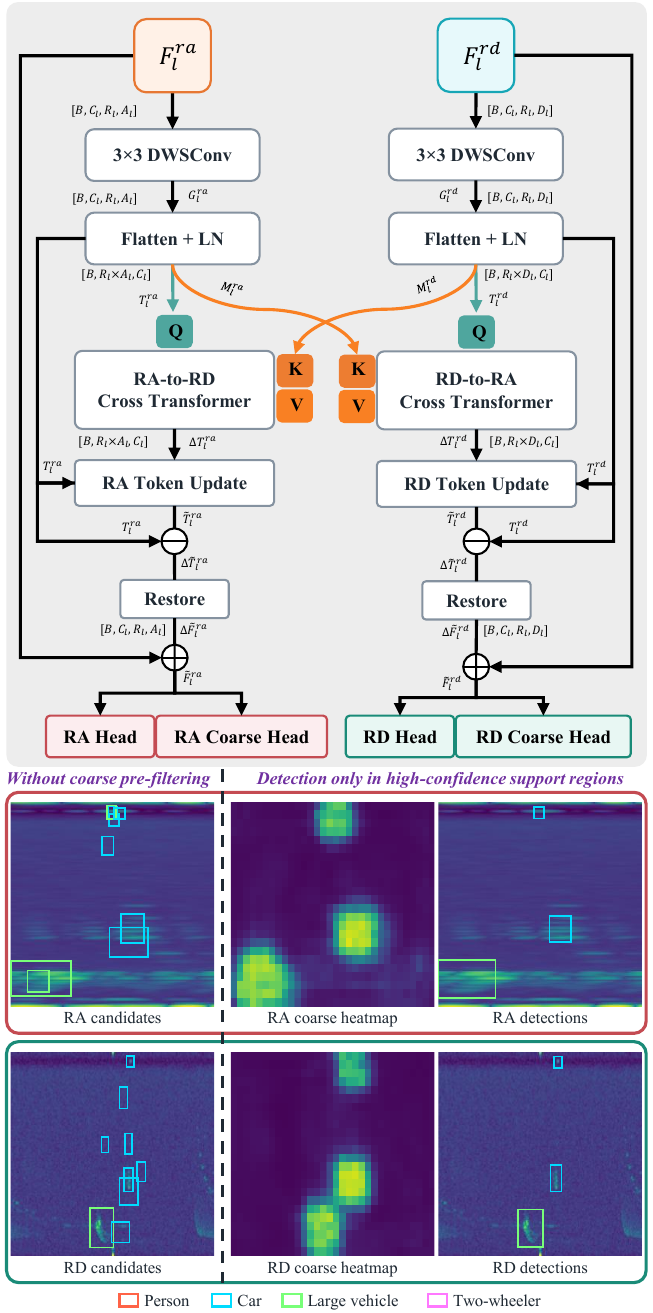}
\caption{Cross-View Auxiliary Supervision. Top: the Cross-View
Transformer performs cross-attention between RA and RD features before
the RA and RD heads. Bottom: coarse object-support pre-filtering.
Without the coarse-support prior, detector candidates accumulate in
unsupported regions and overlap around true object responses; the
predicted support maps suppress such candidates before final decoding
and NMS.}
\label{fig:coarse_support_prefilter}
\vspace{-3mm}
\end{figure}

A lightweight cross-view transformer exchanges information between RA and RD features using pooled memory tokens (Fig.~\ref{fig:coarse_support_prefilter}, top). For example, the RA tokens attend to pooled RD memory:
\begin{equation}
\Delta T^{ra}_{\ell}
=
\operatorname{MHA}
\left(
\operatorname{LN}(T^{ra}_{\ell}),
\operatorname{LN}(M^{rd}_{\ell}),
\operatorname{LN}(M^{rd}_{\ell})
\right),
\end{equation}
where $T^{ra}_{\ell}$ are flattened RA tokens and $M^{rd}_{\ell}$ are pooled RD memory tokens. For either view $v\in\{ra,rd\}$, a $3\times3$ depthwise-separable token-preparation convolution produces $G^{v}_{\ell}$, from which the full-resolution tokens and pooled memory tokens are formed as
\begin{equation}
\begin{aligned}
G^{v}_{\ell}
&=
\operatorname{DWSConv}_{3}\left(F^{v}_{\ell}\right),
\\
T^{v}_{\ell}
&=
\operatorname{Flat}\left(G^{v}_{\ell}\right),
\\
M^{v}_{\ell}
&=
\operatorname{Flat}\left(\operatorname{Pool}_{M}\left(G^{v}_{\ell}\right)\right).
\end{aligned}
\end{equation}
where $\operatorname{Pool}_{M}$ is adaptive average pooling to at most $M\times M$ locations, with $M=12$ in our implementation. The reverse RD-to-RA direction exchanges the roles of the two views. For the RA update, the residual attention and token-MLP stages are

\begin{equation}
\begin{aligned}
\bar{T}^{ra}_{\ell}
&=
T^{ra}_{\ell}
+
\tanh\left(\gamma^{ra}_{\mathrm{attn}}\right)
\Delta T^{ra}_{\ell},\\
\tilde{T}^{ra}_{\ell}
&=
\bar{T}^{ra}_{\ell}
+
\tanh\left(\gamma^{ra}_{\mathrm{ffn}}\right)
\operatorname{MLP}\left(
\operatorname{LN}\left(\bar{T}^{ra}_{\ell}\right)
\right),
\end{aligned}
\end{equation}

where $\gamma^{ra}_{\mathrm{attn}}$ and $\gamma^{ra}_{\mathrm{ffn}}$ are learned residual scales, and the MLP expands the channel dimension from $C$ to $2C$ with GELU before projecting it back to $C$. Finally, the token update is restored to the RA grid and refined before it is added to the original feature:
\begin{equation}
\begin{aligned}
\Delta F^{ra}_{\ell}
&=
\operatorname{Restore}\left(\tilde{T}^{ra}_{\ell}-T^{ra}_{\ell}\right),
\\
\hat{F}^{ra}_{\ell}
&=
F^{ra}_{\ell}
+
\operatorname{Refine}^{ra}_{\ell}\left(\Delta F^{ra}_{\ell}\right).
\end{aligned}
\end{equation}
where, $\operatorname{Refine}^{ra}_{\ell}$ contains two $3\times3$ depthwise-separable convolutional layers. The RD update is defined symmetrically by exchanging $ra$ and $rd$. Thus, spatial RA tokens query compact RD motion memory, while RD tokens query compact RA spatial memory. The CVAS box branches predict 2D RA and RD boxes at every pyramid level, which provide additional view-level supervision and encourage the cube backbone to learn features that remain geometrically consistent after projection.

The CVAS detections use the same anchor decoding form as the cube branch, but in 2D. For RA, the center and size are decoded with $(r,a)$ grid coordinates and $\mathcal{A}^{ra}_{\ell}$; for RD, they are decoded with $(r,d)$ grid coordinates and $\mathcal{A}^{rd}_{\ell}$. This keeps the auxiliary heads geometrically tied to the 3D cube anchors while allowing each view to specialize in its own projection.

\textbf{Coarse Object-Support Guidance.}
The second CVAS path predicts coarse object-support heatmaps that
identify regions likely to contain moving-object radar evidence,
providing a coarse-to-fine localization prior without replacing the
RA/RD box heads or the final cube detector. Compared with sparse
anchor labels alone, the heatmap loss supplies denser gradients
around object-support regions and helps the network learn
radar-specific extent patterns, including elongated RA responses and
range--Doppler motion traces.
Fig.~\ref{fig:coarse_support_prefilter} illustrates how this prior
suppresses overlapping candidates in unsupported RA/RD regions
before final decoding.

Let $H^{ra}$ and $H^{rd}$ denote the predicted coarse heatmaps in RA
and RD views. The corresponding targets are generated by projecting
each radar-aware cube annotation onto the two views and filling the
projected object support with a soft confidence mask. The heatmaps
encode object visibility in each projection, while the final cube
box encodes the complete range--azimuth--Doppler extent. During
training, the CVAS heatmap loss regularizes the auxiliary branches
and improves feature alignment between RA, RD, and RAD
representations. During inference, each heatmap modulates the dense
predictions of its corresponding RA or RD detection branch before
view-wise decoding and NMS. The projected RA/RD support can
additionally gate cube proposals that lack consistent spatial or
Doppler-view evidence.

\subsection{Loss Functions}
\label{sec:loss_functions}

During joint training, RaMOD uses the following multi-task objective:
\begin{equation}
\mathcal{L}
=
\mathcal{L}_{det}^{cube}
+
\lambda_{aux}\mathcal{L}_{det}^{aux}
+
\lambda_{coarse}\mathcal{L}_{coarse}
+
\lambda_{ego}\mathcal{L}_{ego}
+
\lambda_{vel}\mathcal{L}_{vel}.
\end{equation}
The cube detection loss includes box regression, objectness prediction, and category classification in the RAD space. The CVAS detection loss applies the same principle to RA and RD projections. The coarse-location loss supervises the CVAS heatmaps and encourages the model to learn stable object-support regions.

Smooth-$L_1$ regression supervises the ego-motion vector and both
unwrapped and raw object-velocity targets. The Doppler velocity loss
additionally includes binary cross-entropy for wrapping classification:

\begin{equation}
\begin{aligned}
\mathcal{L}_{vel}
=&
\mathcal{L}_{reg}
\left(
\hat{\mathbf{v}}^{\mathrm{unw}},
\mathbf{v}^{\mathrm{unw}}
\right)
+
\mathcal{L}_{reg}
\left(
\hat{\mathbf{v}}^{\mathrm{raw}},
\mathbf{v}^{\mathrm{raw}}
\right)
\\
&+
\lambda_{\mathrm{wrap}}
\mathcal{L}_{\mathrm{bce}}
\left(
\hat{p}_{\mathrm{wrap}},
q_{\mathrm{wrap}}
\right),
\end{aligned}
\end{equation}
where $q_{\mathrm{wrap}}\in\{0,1\}$ is the ground-truth Doppler-wrapping label.

All velocity losses are applied only to positive cube anchors with
valid velocity annotations. During subsequent velocity-head
fine-tuning, the objective is restricted to
$\lambda_{\mathrm{vel}}\mathcal{L}_{\mathrm{vel}}$.

\section{Experiments}
\label{sec:experiments}

\subsection{Experimental Setup}
\label{sec:exp_setup}

\subsubsection{BRIDGE-Liv Dataset and Annotation}

\begin{figure}[!t]
\centering
\includegraphics[width=0.85\linewidth]{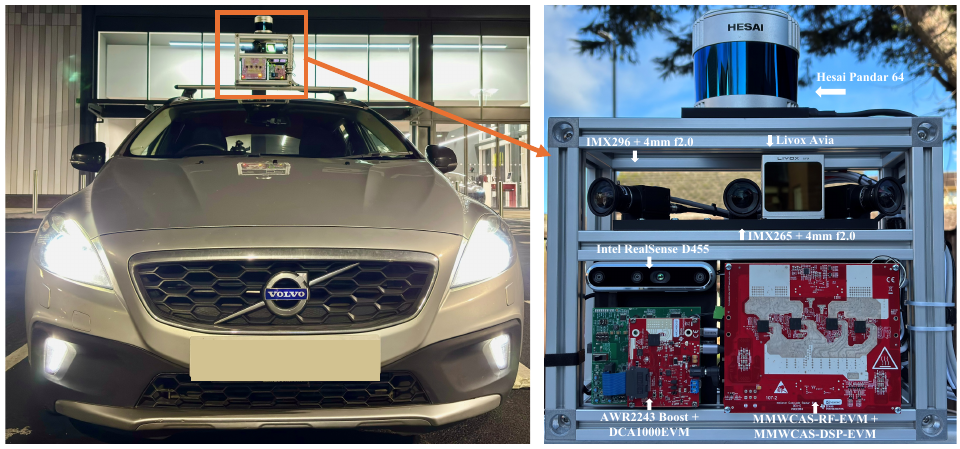}
\caption{The synchronized multi-sensor vehicle platform used for
collecting the BRIDGE-Liv dataset.}
\label{fig:mmliv_platform}
\vspace{-3mm}
\end{figure}

Joint evaluation requires moving-ego RAD cubes with object-level
measurement-extent boxes, velocity labels, and Doppler-wrapping
annotations, a combination that no public benchmark currently
provides. CRUW~\cite{wang2021rodnet} releases only four chirps per
frame, so no Doppler spectrum can be formed. RADIal~\cite{rebut2022radial}
provides DDM ADC data whose RAD reconstruction requires separating the
Doppler-shifted transmitter responses, and its annotations contain
neither RAD boxes nor wrapping labels. K-Radar~\cite{paek2022kradar}
and RaDelft~\cite{roldan2024radelft} natively cover only about
$\pm1.9$ and $\pm2.5$~m/s, so the static field wraps repeatedly under
ego motion (Sec.~\ref{sec:world_static_doppler_field}), and their
physical-space boxes and occupancy labels do not specify RAD
measurement extents. We therefore collected BRIDGE-Liv.

BRIDGE-Liv is collected using the hardware-synchronized vehicle
platform in Fig.~\ref{fig:mmliv_platform}, which integrates two
LiDARs, an IMU, three RGB cameras, RTK, and single-chip and cascaded imaging radars. RaMOD uses only the single-chip AWR2243BOOST,
selected for its wider native Doppler coverage under our radar
configurations. Its two-transmitter, four-receiver configuration
provides a Doppler-bin spacing of $0.182$~m/s and an unambiguous
velocity interval of approximately $[-23.19,23.19)$~m/s.
The radar operates at $10$~Hz, and each cube contains
$192\times256\times255$ bins along the range, azimuth, and Doppler
dimensions, respectively. BRIDGE-Liv contains $76{,}060$
synchronized radar frames, of which $51{,}833$ are used in this study.

Ground-truth annotations are generated using a semi-automatic
multi-sensor pipeline. Camera detections and temporal tracks provide
semantic classes and object identities, while LiDAR geometry and
moving-point information distinguish moving road users from static
movable objects such as parked vehicles. Ego odometry and velocity
are obtained from LiDAR--IMU SLAM. Calibrated sensor extrinsics map
moving-object candidates into radar coordinates for cube-level box
and object-velocity annotation. During annotation, radar visibility
is checked in the cube after applying global static-consistent
Doppler suppression
(Sec.~\ref{sec:world_static_doppler_field}).
All annotations are manually inspected and corrected for projection
errors, closely spaced objects, and tracking failures. All compared
methods are evaluated using the same resulting annotations.
Further platform, calibration, and annotation details will be
documented and released separately.

\subsubsection{Training and Evaluation}

We adopt a fixed sequence-level split of the $51{,}833$ frames used in
this study, with disjoint training and test sequences. The test set
contains $10{,}367$ radar frames. Evaluation considers four 
road-user classes: \emph{person}, \emph{car},
\emph{large vehicle}, and \emph{two-wheeler}. All compared methods use
the same data split, class taxonomy, and ground-truth annotations.

Unless stated otherwise, RaMOD receives $T=3$ consecutive RAD cubes.
The temporal encoder uses a base width of $32$, while each CVAS
Cross-View Transformer uses four attention heads and a memory size of
$M=12$; the coarse object-support term is weighted
by $\lambda_{\mathrm{coarse}}=0.05$. Training proceeds in two stages. The full network is first jointly optimized for $10$ epochs using the multi-task objective. The selected
checkpoint is then refined for $20$ epochs using only the velocity
objective, with all other network components frozen and only the
Velocity Heads updated.

\subsubsection{Evaluation Criteria}
Detection is evaluated by mean average precision (mAP) at IoU thresholds of $0.10$, $0.30$, and $0.50$. The $0.30$ operating point is used for class-wise and ablation analyses, whereas all three thresholds are reported for state-of-the-art comparison. RAD cube mAP uses 3D IoU in range--azimuth--Doppler space; RA and RD mAP use 2D IoU in the corresponding projection. RaMOD cube predictions are projected only for common-view comparison. Ego-State MAE averages absolute errors over evaluated samples
and all six ego-motion components, with translational velocities
expressed in m/s and angular velocities in rad/s. For object-level motion, Velocity MAE denotes the mean of the radial- and forward-velocity MAEs computed from the selected estimate $\hat{\mathbf{v}}$. Doppler-wrapping accuracy compares $\mathbb{I}[\hat{p}_{\mathrm{wrap}}\geq0.5]$ with $q_{\mathrm{wrap}}$. Unless specified otherwise, runtime is measured over the complete evaluation loop.

\subsection{Comparison With State-of-the-Art Methods}
\label{sec:sota_comparison}

\begin{table*}[!t]
\centering
\caption{Detection performance and model complexity on BRIDGE-Liv.
mAP (\%) is reported at IoU $0.10$, $0.30$, and $0.50$.
Params./FLOPs are given in millions/billions.
FPS denotes model-only FP32 throughput at batch size $1$ on an
RTX~5090; a dash indicates no prediction in that space.}
\label{tab:sota_detection}
\scriptsize
\setlength{\tabcolsep}{1.5pt}
\renewcommand{\arraystretch}{1.14}

\begin{tabularx}{\textwidth}{
@{}
>{\raggedright\arraybackslash}p{0.18\textwidth}
>{\centering\arraybackslash}p{0.065\textwidth}
>{\centering\arraybackslash}p{0.105\textwidth}
|YYY|YYY|YYY|
>{\centering\arraybackslash}p{0.045\textwidth}
@{}}

\toprule
\multirow{2}{*}{\textbf{Method}}
&
\multirow{2}{*}{\shortstack{\textbf{Pred.}\\\textbf{Space}}}
&
\multirow{2}{*}{\shortstack{\textbf{Params./FLOPs}\\\textbf{(M/G)}}}
&
\multicolumn{3}{c|}{\textbf{Cube mAP}}
&
\multicolumn{3}{c|}{\textbf{RA mAP}}
&
\multicolumn{3}{c|}{\textbf{RD mAP}}
&
\multirow{2}{*}{\textbf{FPS}}
\\
\cmidrule(lr){4-6}
\cmidrule(lr){7-9}
\cmidrule(lr){10-12}
& & &
\textbf{IoU=0.1} & \textbf{IoU=0.3} & \textbf{IoU=0.5} &
\textbf{IoU=0.1} & \textbf{IoU=0.3} & \textbf{IoU=0.5} &
\textbf{IoU=0.1} & \textbf{IoU=0.3} & \textbf{IoU=0.5} &
\\
\midrule

RODNet~\cite{wang2021rodnet}
& RA & 7.34/59.91
& -- & -- & --
& 33.5 & 31.7 & 21.1
& -- & -- & --
& 458.7 \\

T-RODNet~\cite{jiang2023trodnet}
& RA & 32.31/409.79
& -- & -- & --
& 28.3 & 23.9 & 12.8
& -- & -- & --
& 89.7 \\

FFT-RadNet~\cite{rebut2022radial}
& RA & 2.65/12.79
& -- & -- & --
& 30.3 & 26.6 & 16.0
& -- & -- & --
& 283.6 \\

T-FFTRadNet~\cite{giroux2023tfftradnet}
& RA & 2.32/5.23
& -- & -- & --
& 34.5 & 28.2 & 14.1
& -- & -- & --
& 247.5 \\

RadarFormer~\cite{dalbah2023radarformer}
& RA & 1.95/368.62
& -- & -- & --
& 29.8 & 24.7 & 13.3
& -- & -- & --
& 15.9 \\

\midrule

MV-Net~\cite{ouaknine2021mvnet}
& RA/RD & 3.32/109.13
& -- & -- & --
& 26.2 & 19.0 & 8.1
& 75.2 & 74.1 & 61.1
& 232.7 \\

TMVA-Net~\cite{ouaknine2021mvnet}
& RA/RD & 5.69/225.46
& -- & -- & --
& 36.1 & 28.9 & 14.1
& 57.5 & 56.6 & 48.1
& 141.1 \\

PeakConv~\cite{zhang2023peakconv}
& RA/RD & 6.38/257.30
& -- & -- & --
& 27.1 & 23.5 & 12.4
& 61.3 & 59.2 & 39.8
& 44.9 \\

AdaPKC~\cite{li2024adapkc}
& RA/RD & 6.38/257.30
& -- & -- & --
& 32.2 & 25.5 & 10.8
& 63.0 & 61.6 & 50.9
& 19.3 \\

\midrule

RADDet~\cite{zhang2021raddet}
& RAD cube & 9.90/31.93
& 68.1 & 57.0 & 24.0
& 69.4 & 64.1 & 43.7
& 71.1 & 67.8 & 52.6
& 192.0 \\

Mamba-RODNet~\cite{zhuang2025rawadc}
& RAD cube & 20.38/100.23
& 75.4 & 65.1 & 23.7
& 77.4 & 72.4 & 46.8
& 77.0 & 73.4 & 57.5
& 64.6 \\

\rowcolor{rdoBest}
\textbf{RaMOD (ours)}
& RAD cube & \textbf{11.39/149.69}
& \best{84.4} & \best{80.8} & \best{56.0}
& \best{84.5} & \best{82.4} & \best{71.6}
& \best{85.0} & \best{84.8} & \best{81.5}
& 107.5 \\

\bottomrule
\end{tabularx}
\vspace{-2mm}
\end{table*}

We compare representative raw-radar detectors spanning RA, RA/RD, and
RAD cube output spaces under the same BRIDGE-Liv protocol
(Table~\ref{tab:sota_detection}). Confidence-map and segmentation
baselines are adapted with RA or RA/RD box-detection heads and trained
with box-level supervision. PeakConv~\cite{zhang2023peakconv} and
AdaPKC~\cite{li2024adapkc} are adapted to the RA/RD setting, while
Mamba-RODNet is adapted from the raw-ADC formulation
of~\cite{zhuang2025rawadc} to produce RAD cube detections. RaMOD cube
detections are projected to RA and RD for common-view comparison; cube
mAP is reported only for methods that output 3D RAD boxes.

\subsubsection{Object Detection}
RaMOD attains $84.4\%$, $80.8\%$, and $56.0\%$ cube mAP at IoU $0.10$,
$0.30$, and $0.50$, outperforming Mamba-RODNet, the strongest RAD cube
baseline, by $9.0$, $15.7$, and $32.3$ percentage points, respectively.
The widening margin under stricter IoU criteria indicates more accurate
RAD localization and supports retaining the RAD representation together
with explicit ego-motion-aware Doppler reasoning. At IoU $0.30$, RaMOD
also exceeds Mamba-RODNet in projected RA and RD mAP by $10.0$ and $11.4$
points while providing explicit RAD extents and object-level motion
outputs.

Table~\ref{tab:per_class_cube} shows consistent gains over Mamba-RODNet across all classes: $16.5$ points for persons, $18.8$ for cars, $18.5$ for large vehicles, and $9.1$ for two-wheelers. The person gain indicates that the multi-scale representation retains weak and compact responses, while the gains for cars and large vehicles show that RaMOD exploits extended spatial support and distributed Doppler evidence rather than a single RA peak.

\begin{table}[!t]
\centering
\caption{Per-class cube AP (\%) at IoU $0.30$.}
\label{tab:per_class_cube}
\scriptsize
\setlength{\tabcolsep}{1.5pt}
\renewcommand{\arraystretch}{1.14}
\begin{tabularx}{\columnwidth}{@{}>{\raggedright\arraybackslash}p{0.25\columnwidth}>{\centering\arraybackslash}p{0.08\columnwidth}>{\centering\arraybackslash}p{0.12\columnwidth}>{\centering\arraybackslash}p{0.10\columnwidth}>{\centering\arraybackslash}p{0.17\columnwidth}>{\centering\arraybackslash}p{0.20\columnwidth}@{}}
\toprule
\textbf{Method} & \textbf{mAP} & \textbf{Person} & \textbf{Car} & \textbf{Large Veh.} & \textbf{Two-Wheeler} \\
\midrule
RADDet & 57.0 & 50.4 & 75.4 & 72.5 & 29.7 \\
Mamba-RODNet & 65.1 & 50.4 & 74.7 & 78.8 & 56.5 \\
\rowcolor{rdoBest}\textbf{RaMOD (ours)} & \best{80.8} & \best{66.9} & \best{93.5} & \best{97.3} & \best{65.6} \\
\bottomrule
\end{tabularx}
\vspace{-3mm}
\end{table}

\subsubsection{Object-Level Velocity Estimation}
Most compared baselines are designed for detection only and do not provide object-associated velocity estimates or Doppler-wrapping decisions, precluding a direct velocity comparison. We therefore evaluate the final velocity-refined RaMOD checkpoint on the held-out test split. It achieves velocity, radial-velocity, and forward-velocity MAEs of $1.128$, $1.094$, and $1.161~\mathrm{m/s}$, respectively, with $99.87\%$ Doppler-wrapping accuracy.

\subsubsection{Qualitative Results}
\begin{figure*}[!t]
\centering
\includegraphics[width=\textwidth]{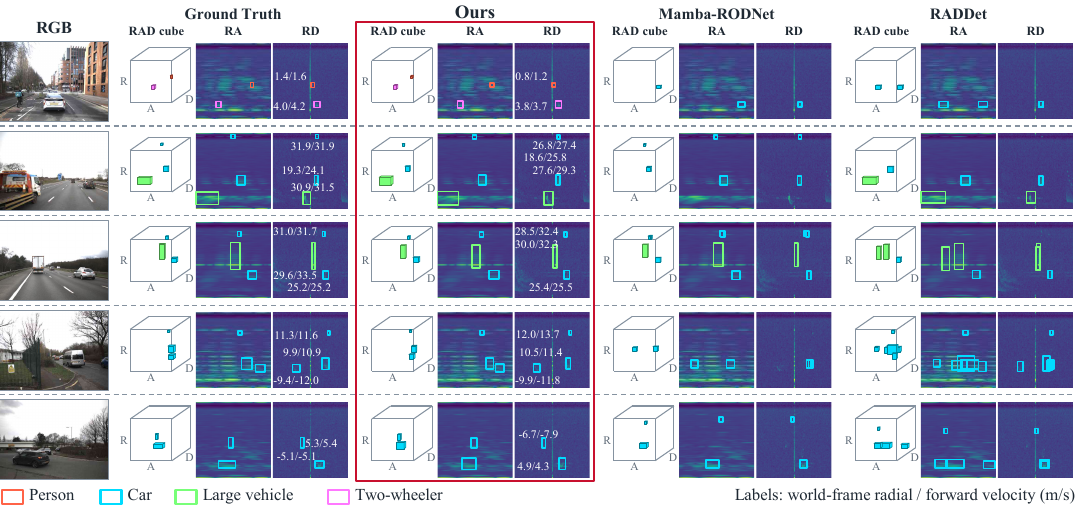}
\caption{Qualitative RAD cube results with RGB references, annotations, and representative cube-detector predictions. RaMOD preserves spatial and Doppler extents with category and velocity labels.}
\label{fig:qualitative_rad_cube}
\vspace{-1mm}
\end{figure*}

Figure~\ref{fig:qualitative_rad_cube} shows that RaMOD preserves each
object's range--azimuth--Doppler extent, category, and
velocity labels in urban and highway scenes.

\begin{figure*}[!t]
\centering
\includegraphics[width=\textwidth]{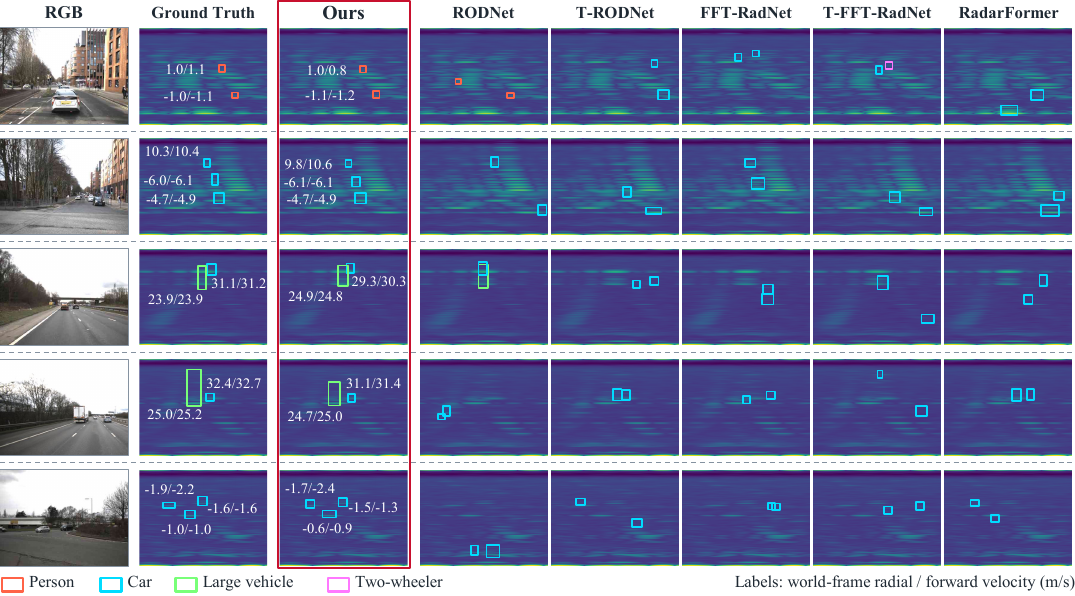}
\caption{Qualitative RA comparison with RGB references, annotations,
RA-based methods, and RaMOD. RaMOD maintains consistent projected
range--azimuth localization in cluttered and multi-object scenes.}
\label{fig:qualitative_ra}
\vspace{-5mm}
\end{figure*}

Figure~\ref{fig:qualitative_ra} shows consistent projected RA
localization for compact road users and extended vehicles despite
strong static responses and extended target signatures.

\begin{figure*}[!t]
\centering
\includegraphics[width=0.93\textwidth]{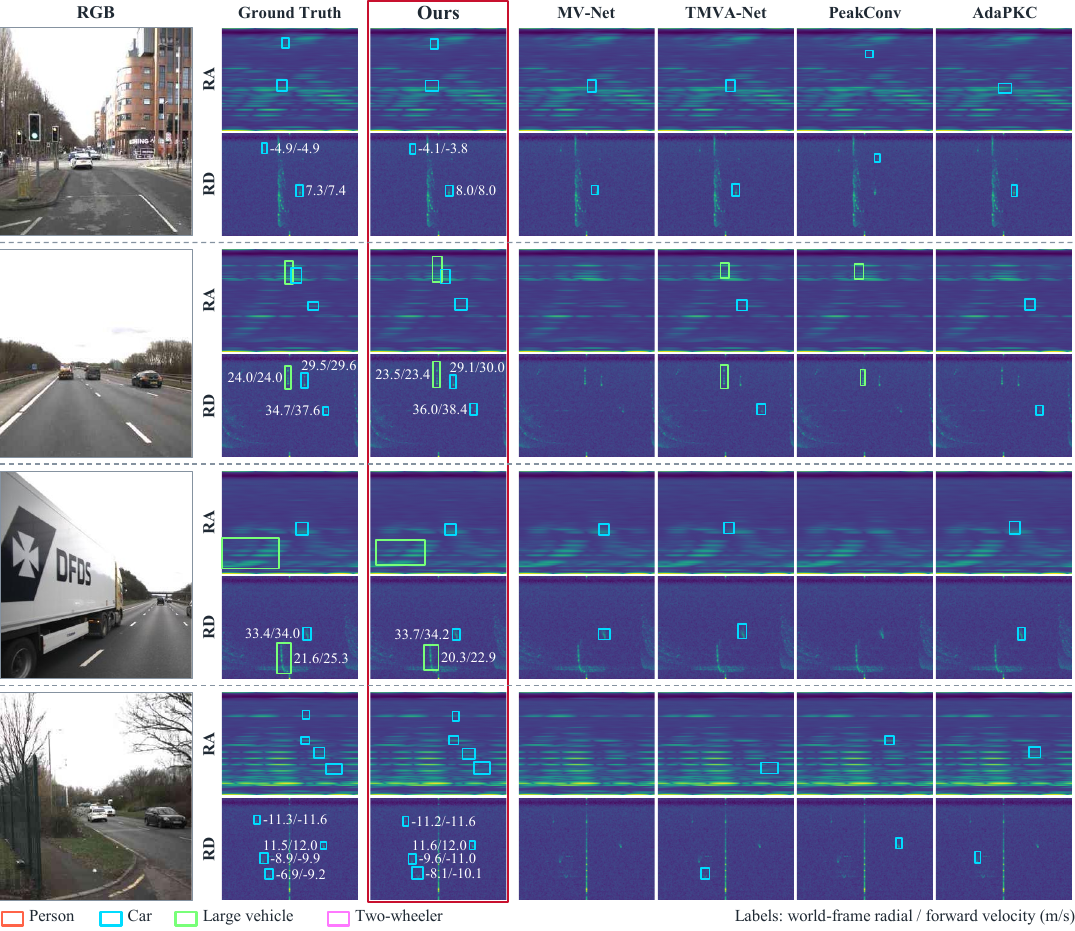}
\caption{Qualitative RA/RD comparison with RGB references and detections from multi-view methods and RaMOD.}
\label{fig:qualitative_ra_rd}
\vspace{-2mm}
\end{figure*}

Figure~\ref{fig:qualitative_ra_rd} shows that RaMOD recovers target support
in both views under ego-induced static patterns and projection ambiguity
while retaining a coherent RAD cube hypothesis.

\subsection{Ablation and Sensitivity Analysis}
\label{sec:component_ablation}
We evaluate the core components and temporal-window length under matched first-stage training protocols. Unless otherwise stated, all variants use $\lambda_{\mathrm{coarse}}=0.05$, $T=3$, and the same 10-epoch schedule. Table~\ref{tab:component_ablation} reports cube detection, ego-state, and object-level velocity metrics. Cube mAP at IoU $0.30$ is the primary ablation criterion; the remaining metrics diagnose the corresponding stage-1 outputs and need not vary monotonically with detection accuracy.

\subsubsection{Ego-Motion Supervision}
Removing ego supervision causes the largest detection degradation among the individual representation-learning components: cube mAP at IoU $0.30$ drops by $6.69$ points, from $80.85\%$ to $74.16\%$, while Ego-State MAE increases from $0.129$ to $2.755$. This result shows that explicit ego supervision is required to learn a reliable static-consistent Doppler prior and separate independently moving returns from the moving-ego background, rather than merely providing an auxiliary output.

\subsubsection{Cross-Scale Feature Refinement}
Disabling CS-ARF reduces cube mAP at IoU $0.30$ by $3.35$ points, demonstrating the importance of explicit information exchange among the $C_3$, $C_4$, and $C_5$ RAD feature scales before pyramid decoding. Cross-scale refinement improves the representation of weak, extended, and cluttered radar responses.

\subsubsection{Cross-View and Coarse-Support Components}
\begin{figure}[!t]
\centering
\includegraphics[width=0.88\columnwidth]{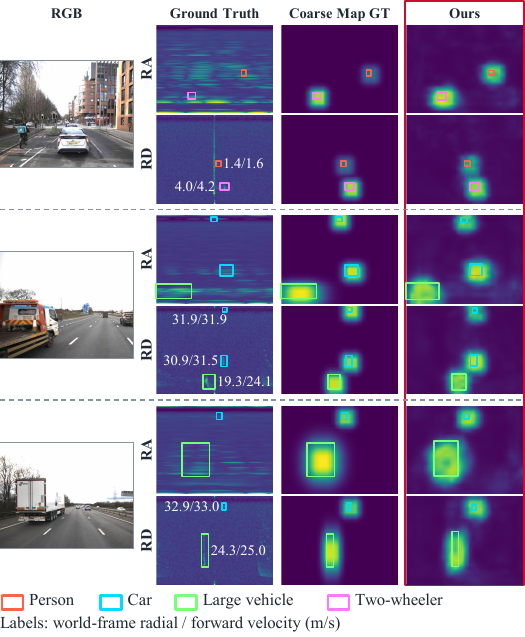}
\caption{Coarse object-support guidance in the RA and RD views. For each scene, the raw projection is followed by the target support and predicted coarse heatmaps. The predicted support highlights target-related regions and provides the pre-filtering prior for view-wise candidate decoding.}
\label{fig:qualitative_coarse_support}
\vspace{-3mm}
\end{figure}

The CVAS branches provide RA/RD supervision, while coarse object-support maps guide view-wise candidate decoding. Replacing the Cross-View Transformer with convolutional cross-view interaction lowers cube mAP at IoU $0.30$ by $5.26$ points, and removing RA/RD auxiliary supervision causes a $1.87$-point decline. Removing coarse-support guidance lowers mAP by $3.62$ points, showing that the support maps provide a coarse localization prior that filters unsupported RA/RD candidates before fine box decoding. Figure~\ref{fig:qualitative_coarse_support} provides qualitative evidence: the predicted heatmaps concentrate on target-related regions and exclude broad areas that otherwise admit spurious or duplicate candidates.

\subsubsection{Velocity and Doppler-Wrap Learning}
Without the object-level Velocity Head, cube mAP at IoU $0.30$ decreases by $5.20$ points, indicating that motion regression provides useful supervision to the shared cube representation. Removing the Doppler-wrap loss reduces wrapping accuracy from $99.86\%$ to $58.42\%$ and cube mAP from $80.85\%$ to $75.39\%$. This result confirms that explicit wrapping supervision is essential for resolving Doppler ambiguity.

\begin{table}[!t]
\centering
\caption{Core component ablation under the matched $\lambda_{\mathrm{coarse}}=0.05$, $T=3$, and 10-epoch detector protocol. Detection entries are cube mAP (\%) at IoU $0.30$. ``--'' denotes that no velocity output is predicted.}
\label{tab:component_ablation}
\scriptsize
\setlength{\tabcolsep}{1.2pt}
\renewcommand{\arraystretch}{1.14}
\resizebox{\columnwidth}{!}{%
\begin{tabular}{@{}>{\raggedright\arraybackslash}p{0.29\columnwidth}>{\centering\arraybackslash}p{0.19\columnwidth}>{\centering\arraybackslash}p{0.14\columnwidth}>{\centering\arraybackslash}p{0.18\columnwidth}>{\centering\arraybackslash}p{0.15\columnwidth}@{}}
\toprule
\textbf{Variant} & \textbf{Cube mAP (\%)} & \textbf{Ego-State MAE} & \textbf{Velocity MAE (m/s)} & \textbf{Wrap Acc. (\%)} \\
\midrule
\rowcolor{rdoBest}\textbf{RaMOD (stage 1)} & \best{80.85} & 0.129 & 1.465 & \best{99.86} \\
w/o ego supervision & 74.16 & 2.755 & 1.782 & 99.83 \\
w/o Cross-View Transformer & 75.59 & 0.115 & 1.215 & 99.79 \\
w/o CS-ARF & 77.50 & 0.109 & 1.335 & 99.80 \\
w/o RA/RD auxiliary supervision & 78.98 & 0.165 & 1.546 & 99.81 \\
w/o coarse support & 77.23 & 0.135 & 1.376 & 99.80 \\
w/o Velocity Head & 75.65 & 0.139 & -- & -- \\
w/o wrap loss & 75.39 & 0.126 & 1.415 & 58.42 \\
\bottomrule
\end{tabular}}
\vspace{-5mm}
\end{table}

\subsubsection{Temporal-Window Sensitivity}
\label{sec:parameter_sensitivity}

Under the same final configuration, cube mAP reaches $75.76\%$,
$80.85\%$, and $78.34\%$ for $T=1$, $3$, and $5$, respectively.
The three-frame window improves mAP by $5.09$ points over a single
cube, demonstrating the value of short-term temporal evidence.
Although $T=5$ improves over $T=1$, its mAP remains $2.51$ points
below $T=3$ with a larger input. We therefore use $T=3$ by default.

\subsection{Model Complexity and Runtime}
\label{sec:model_complexity}
Table~\ref{tab:sota_detection} reports model-only complexity and
runtime, excluding data loading, decoding, NMS, and metric accumulation.
All networks are evaluated using one $T=3$ RAD sequence per FP32 forward
pass on the RTX~5090. FLOPs count the executed convolution, linear, and
multi-head-attention operations, with one multiply-add counted as two
FLOPs; normalization, pooling, elementwise operations, and custom
indexing are excluded.

RaMOD uses fewer parameters than T-RODNet and Mamba-RODNet while
processing the 3D RAD representation. Despite its $149.69$~G FLOPs, it
achieves $107.5$ model-only FPS. The complete evaluation loop runs at
$25.7$ FPS, including data loading, network inference, cube decoding,
NMS, and per-frame metric accumulation, exceeding the $10$~Hz radar
sensing rate and supporting real-time operation.

\section{Conclusion and Future Work}

In this paper, we presented RaMOD, a real-time moving-object detection
and velocity estimation framework that directly processes one or more
consecutive raw RAD cubes from a moving single-chip mmWave radar.
RaMOD combines ego-motion-guided static-consistent Doppler suppression
with temporal multi-scale cube encoding to detect and classify moving
objects in RAD space. It estimates frame-wise ego motion and
object-level radial and forward velocities, and predicts
Doppler-wrapping states with unwrapped velocity estimates.
Experiments on BRIDGE-Liv demonstrate that RaMOD outperforms all
compared raw-radar methods across the RAD-cube, RA, and RD detection
spaces at all reported IoU thresholds, while maintaining accurate
motion estimation and real-time operation. The RaMOD implementation
and BRIDGE-Liv dataset will be open-sourced.

Future work will extend RaMOD to motion-aware 3D occupancy of static
structures and moving road users. Joint estimation of object heading
and planar velocity will relax the current ego-forward motion
assumption, and an azimuth-dependent suppression region that follows
the static field at each azimuth will be evaluated together with the
corresponding annotation update. In parallel, we are establishing a
DDM acquisition platform to extend validation beyond the current
TDM-MIMO configuration, including ego-motion-induced static Doppler
and velocity ambiguity. As BRIDGE-Liv expands, we will explore more
efficient, data-scalable end-to-end radar models and their potential
integration into multimodal VLA driving systems.

\bibliographystyle{IEEEtran}
\bibliography{ref}

\begin{thebibliography}{10}
\providecommand{\url}[1]{#1}
\csname url@samestyle\endcsname
\providecommand{\newblock}{\relax}
\providecommand{\bibinfo}[2]{#2}
\providecommand{\BIBentrySTDinterwordspacing}{\spaceskip=0pt\relax}
\providecommand{\BIBentryALTinterwordstretchfactor}{4}
\providecommand{\BIBentryALTinterwordspacing}{\spaceskip=\fontdimen2\font plus
\BIBentryALTinterwordstretchfactor\fontdimen3\font minus
  \fontdimen4\font\relax}
\providecommand{\BIBforeignlanguage}[2]{{%
\expandafter\ifx\csname l@#1\endcsname\relax
\typeout{** WARNING: IEEEtran.bst: No hyphenation pattern has been}%
\typeout{** loaded for the language `#1'. Using the pattern for}%
\typeout{** the default language instead.}%
\else
\language=\csname l@#1\endcsname
\fi
#2}}
\providecommand{\BIBdecl}{\relax}
\BIBdecl

\bibitem{li2022bevformer}
Z.~Li, W.~Wang, H.~Li, E.~Xie, C.~Sima, T.~Lu, Y.~Qiao, and J.~Dai,
  ``{BEVFormer}: Learning bird's-eye-view representation from multi-camera
  images via spatiotemporal transformers,'' in \emph{Computer Vision -- ECCV
  2022}.\hskip 1em plus 0.5em minus 0.4em\relax Springer, 2022, pp. 1--18.

\bibitem{lang2019pointpillars}
A.~H. Lang, S.~Vora, H.~Caesar, L.~Zhou, J.~Yang, and O.~Beijbom,
  ``{PointPillars}: Fast encoders for object detection from point clouds,'' in
  \emph{Proceedings of the IEEE/CVF Conference on Computer Vision and Pattern
  Recognition (CVPR)}, 2019, pp. 12\,697--12\,705.

\bibitem{Hu_2023_CVPR}
Y.~Hu, J.~Yang, L.~Chen, K.~Li, C.~Sima, X.~Zhu, S.~Chai, S.~Du, T.~Lin,
  W.~Wang, L.~Lu, X.~Jia, Q.~Liu, J.~Dai, Y.~Qiao, and H.~Li,
  ``Planning-oriented autonomous driving,'' in \emph{2023 IEEE/CVF Conference
  on Computer Vision and Pattern Recognition (CVPR)}, 2023, pp.
  17\,853--17\,862.

\bibitem{zhou2025opendrivevla}
X.~Zhou, X.~Han, F.~Yang, Y.~Ma, V.~Tresp, and A.~Knoll, ``{OpenDriveVLA}:
  Towards end-to-end autonomous driving with large vision language action
  model,'' in \emph{Proceedings of the AAAI Conference on Artificial
  Intelligence}, vol.~40, no.~16, 2026, pp. 13\,782--13\,790.

\bibitem{teed2020raft}
Z.~Teed and J.~Deng, ``{RAFT}: Recurrent all-pairs field transforms for optical
  flow,'' in \emph{Computer Vision -- ECCV 2020}.\hskip 1em plus 0.5em minus
  0.4em\relax Springer, 2020, pp. 402--419.

\bibitem{liu2019flownet3d}
X.~Liu, C.~R. Qi, and L.~J. Guibas, ``{FlowNet3D}: Learning scene flow in {3D}
  point clouds,'' in \emph{Proceedings of the IEEE/CVF Conference on Computer
  Vision and Pattern Recognition (CVPR)}, 2019, pp. 529--537.

\bibitem{Bijelic_2020_CVPR}
M.~Bijelic, T.~Gruber, F.~Mannan, F.~Kraus, W.~Ritter, K.~Dietmayer, and
  F.~Heide, ``Seeing through fog without seeing fog: Deep multimodal sensor
  fusion in unseen adverse weather,'' in \emph{2020 IEEE/CVF Conference on
  Computer Vision and Pattern Recognition (CVPR)}.\hskip 1em plus 0.5em minus
  0.4em\relax IEEE, 2020, pp. 11\,679--11\,689.

\bibitem{millimap}
C.~X. Lu, S.~Rosa, P.~Zhao, B.~Wang, C.~Chen, J.~A. Stankovic, N.~Trigoni, and
  A.~Markham, ``See through smoke: robust indoor mapping with low-cost {mmWave}
  radar,'' in \emph{Proceedings of the 18th International Conference on Mobile
  Systems, Applications, and Services}, 2020, pp. 14--27.

\bibitem{1843aop}
{Texas Instruments}, ``{AWR1843AOPEVM}: {AWR1843AOP} evaluation module for
  single-chip 76-{GHz} to 81-{GHz} automotive radar sensor,'' Product page,
  accessed: 2026-08-27.

\bibitem{patole2017automotive}
S.~M. Patole, M.~Torlak, D.~Wang, and M.~Ali, ``Automotive radars: A review of
  signal processing techniques,'' \emph{IEEE Signal Processing Magazine},
  vol.~34, no.~2, pp. 22--35, 2017.

\bibitem{girshick2014rcnn}
R.~Girshick, J.~Donahue, T.~Darrell, and J.~Malik, ``Rich feature hierarchies
  for accurate object detection and semantic segmentation,'' in
  \emph{Proceedings of the IEEE Conference on Computer Vision and Pattern
  Recognition (CVPR)}, 2014, pp. 580--587.

\bibitem{redmon2016yolo}
J.~Redmon, S.~Divvala, R.~Girshick, and A.~Farhadi, ``You only look once:
  Unified, real-time object detection,'' in \emph{2016 IEEE Conference on
  Computer Vision and Pattern Recognition (CVPR)}, 2016, pp. 779--788.

\bibitem{carion2020detr}
N.~Carion, F.~Massa, G.~Synnaeve, N.~Usunier, A.~Kirillov, and S.~Zagoruyko,
  ``End-to-end object detection with transformers,'' in \emph{Computer Vision
  -- ECCV 2020}.\hskip 1em plus 0.5em minus 0.4em\relax Springer, 2020, pp.
  213--229.

\bibitem{wang2022detr3d}
Y.~Wang, V.~C. Guizilini, T.~Zhang, Y.~Wang, H.~Zhao, and J.~Solomon,
  ``{DETR3D}: {3D} object detection from multi-view images via {3D}-to-{2D}
  queries,'' in \emph{Proceedings of the 5th Conference on Robot Learning},
  ser. Proceedings of Machine Learning Research, vol. 164.\hskip 1em plus 0.5em
  minus 0.4em\relax PMLR, 2022, pp. 180--191.

\bibitem{wei2023surroundocc}
Y.~Wei, L.~Zhao, W.~Zheng, Z.~Zhu, J.~Zhou, and J.~Lu, ``{SurroundOcc}:
  Multi-camera {3D} occupancy prediction for autonomous driving,'' in
  \emph{Proceedings of the IEEE/CVF International Conference on Computer Vision
  (ICCV)}, 2023, pp. 21\,729--21\,740.

\bibitem{zhou2018voxelnet}
Y.~Zhou and O.~Tuzel, ``{VoxelNet}: End-to-end learning for point cloud based
  {3D} object detection,'' in \emph{Proceedings of the IEEE Conference on
  Computer Vision and Pattern Recognition (CVPR)}, 2018, pp. 4490--4499.

\bibitem{shi2020pvrcnn}
S.~Shi, C.~Guo, L.~Jiang, Z.~Wang, J.~Shi, X.~Wang, and H.~Li, ``{PV-RCNN}:
  Point-voxel feature set abstraction for {3D} object detection,'' in
  \emph{Proceedings of the IEEE/CVF Conference on Computer Vision and Pattern
  Recognition (CVPR)}, 2020, pp. 10\,529--10\,538.

\bibitem{yin2021centerpoint}
T.~Yin, X.~Zhou, and P.~Krahenbuhl, ``Center-based {3D} object detection and
  tracking,'' in \emph{Proceedings of the IEEE/CVF conference on computer
  vision and pattern recognition}, 2021, pp. 11\,784--11\,793.

\bibitem{zhang2022bytetrack}
Y.~Zhang, P.~Sun, Y.~Jiang, D.~Yu, F.~Weng, Z.~Yuan, P.~Luo, W.~Liu, and
  X.~Wang, ``{ByteTrack}: Multi-object tracking by associating every detection
  box,'' in \emph{Computer Vision -- ECCV 2022}.\hskip 1em plus 0.5em minus
  0.4em\relax Springer, 2022, pp. 1--21.

\bibitem{chen2021movingobjectsegmentation}
X.~Chen, S.~Li, B.~Mersch, L.~Wiesmann, J.~Gall, J.~Behley, and C.~Stachniss,
  ``Moving object segmentation in {3D} {LiDAR} data: A learning-based approach
  exploiting sequential data,'' \emph{IEEE Robotics and Automation Letters},
  vol.~6, no.~4, pp. 6529--6536, 2021.

\bibitem{wu2024mdetector}
H.~Wu, Y.~Li, W.~Xu, F.~Kong, and F.~Zhang, ``Moving event detection from
  {LiDAR} point streams,'' \emph{Nature Communications}, vol.~15, no.~1, p.
  345, 2024.

\bibitem{danzer2019pointnet}
A.~Danzer, T.~Griebel, M.~Bach, and K.~Dietmayer, ``{2D} car detection in radar
  data with {PointNets},'' in \emph{2019 IEEE Intelligent Transportation
  Systems Conference (ITSC)}, 2019, pp. 61--66.

\bibitem{scheiner2021radarpc}
N.~Scheiner, F.~Kraus, N.~Appenrodt, J.~Dickmann, and B.~Sick, ``Object
  detection for automotive radar point clouds---a comparison,'' \emph{AI
  Perspectives}, vol.~3, no.~1, p.~6, 2021.

\bibitem{palffy2022vod}
A.~Palffy, E.~Pool, S.~Baratam, J.~F.~P. Kooij, and D.~M. Gavrila,
  ``Multi-class road user detection with {3+1D} radar in the view-of-delft
  dataset,'' \emph{IEEE Robotics and Automation Letters}, vol.~7, no.~2, pp.
  4961--4968, 2022.

\bibitem{xu2021rpfa}
B.~Xu, X.~Zhang, L.~Wang, X.~Hu, Z.~Li, S.~Pan, J.~Li, and Y.~Deng,
  ``{RPFA-Net}: A {4D} radar pillar feature attention network for {3D} object
  detection,'' in \emph{2021 IEEE International Intelligent Transportation
  Systems Conference (ITSC)}, 2021, pp. 3061--3066.

\bibitem{liu2023smurf}
J.~Liu, Q.~Zhao, W.~Xiong, T.~Huang, Q.-L. Han, and B.~Zhu, ``{SMURF}: Spatial
  multi-representation fusion for {3D} object detection with {4D} imaging
  radar,'' \emph{IEEE Transactions on Intelligent Vehicles}, vol.~9, no.~1, pp.
  799--812, 2024.

\bibitem{musiat2024radarpillars}
A.~Musiat, L.~Reichardt, M.~Schulze, and O.~Wasenm{\"u}ller, ``{RadarPillars}:
  Efficient object detection from {4D} radar point clouds,'' in \emph{2024 IEEE
  27th International Conference on Intelligent Transportation Systems (ITSC)},
  2024, pp. 1656--1663.

\bibitem{kellner2013instantaneous}
D.~Kellner, M.~Barjenbruch, J.~Klappstein, J.~Dickmann, and K.~Dietmayer,
  ``Instantaneous ego-motion estimation using {Doppler} radar,'' in \emph{16th
  International IEEE Conference on Intelligent Transportation Systems (ITSC
  2013)}, 2013, pp. 869--874.

\bibitem{rapp2017probabilistic}
M.~Rapp, M.~Barjenbruch, M.~Hahn, J.~Dickmann, and K.~Dietmayer,
  ``Probabilistic ego-motion estimation using multiple automotive radar
  sensors,'' \emph{Robotics and Autonomous Systems}, vol.~89, pp. 136--146,
  2017.

\bibitem{zhu2023deepego}
S.~Zhu, A.~Yarovoy, and F.~Fioranelli, ``{DeepEgo}: Deep instantaneous
  ego-motion estimation using automotive radar,'' \emph{IEEE Transactions on
  Radar Systems}, vol.~1, pp. 166--180, 2023.

\bibitem{sie2024radarize}
E.~Sie, X.~Wu, H.~Guo, and D.~Vasisht, ``{Radarize}: Enhancing radar {SLAM}
  with generalizable {Doppler}-based odometry,'' in \emph{Proceedings of the
  22nd Annual International Conference on Mobile Systems, Applications and
  Services}.\hskip 1em plus 0.5em minus 0.4em\relax ACM, 2024, pp. 331--344.

\bibitem{ouaknine2021carrada}
A.~Ouaknine, A.~Newson, J.~Rebut, F.~Tupin, and P.~P{\'e}rez, ``{CARRADA}
  dataset: Camera and automotive radar with range-angle-{Doppler}
  annotations,'' in \emph{2020 25th International Conference on Pattern
  Recognition (ICPR)}, 2021, pp. 5068--5075.

\bibitem{ouaknine2021mvnet}
A.~Ouaknine, A.~Newson, P.~P{\'e}rez, F.~Tupin, and J.~Rebut, ``Multi-view
  radar semantic segmentation,'' in \emph{2021 IEEE/CVF International
  Conference on Computer Vision (ICCV)}, 2021, pp. 15\,651--15\,660.

\bibitem{zhang2023peakconv}
L.~Zhang, X.~Zhang, Y.~Zhang, Y.~Guo, Y.~Chen, X.~Huang, and Z.~Ma,
  ``{PeakConv}: Learning peak receptive field for radar semantic
  segmentation,'' in \emph{2023 IEEE/CVF Conference on Computer Vision and
  Pattern Recognition (CVPR)}, 2023, pp. 17\,577--17\,586.

\bibitem{li2024adapkc}
T.~Li, L.~Zhang, Y.~Zhang, Z.~Hu, P.~Pi, Z.~Lu, Q.~Liao, and Z.~Ma, ``{AdaPKC}:
  {PeakConv} with adaptive peak receptive field for radar semantic
  segmentation,'' in \emph{Advances in Neural Information Processing Systems},
  vol.~37, 2024, pp. 136\,545--136\,575.

\bibitem{chen2026marss}
F.~Chen, T.~Tan, T.~Li, Y.~Quan, and Q.~Liao, ``{MARSS}: Radar semantic
  segmentation via modular attention and state space models,'' in
  \emph{Proceedings of the IEEE/CVF Conference on Computer Vision and Pattern
  Recognition (CVPR)}, 2026, pp. 17\,959--17\,968.

\bibitem{decourt2022darod}
C.~Decourt, R.~VanRullen, D.~Salle, and T.~Oberlin, ``{DAROD}: A deep
  automotive radar object detector on range-{Doppler} maps,'' in \emph{2022
  IEEE Intelligent Vehicles Symposium (IV)}, 2022, pp. 112--118.

\bibitem{gao2021rampcnn}
X.~Gao, G.~Xing, S.~Roy, and H.~Liu, ``{RAMP-CNN}: A novel neural network for
  enhanced automotive radar object recognition,'' \emph{IEEE Sensors Journal},
  vol.~21, no.~4, pp. 5119--5132, 2021.

\bibitem{wang2021rodnet}
Y.~Wang, Z.~Jiang, Y.~Li, J.-N. Hwang, G.~Xing, and H.~Liu, ``{RODNet}: A
  real-time radar object detection network cross-supervised by camera-radar
  fused object {3D} localization,'' \emph{IEEE Journal of Selected Topics in
  Signal Processing}, vol.~15, no.~4, pp. 954--967, 2021.

\bibitem{jiang2023trodnet}
T.~Jiang, L.~Zhuang, Q.~An, J.~Wang, K.~Xiao, and A.~Wang, ``{T-RODNet}:
  Transformer for vehicular millimeter-wave radar object detection,''
  \emph{IEEE Transactions on Instrumentation and Measurement}, vol.~72, pp.
  1--12, 2023.

\bibitem{decourt2024record}
C.~Decourt, R.~VanRullen, D.~Salle, and T.~Oberlin, ``A recurrent {CNN} for
  online object detection on raw radar frames,'' \emph{IEEE Transactions on
  Intelligent Transportation Systems}, vol.~25, no.~10, pp. 13\,432--13\,441,
  2024.

\bibitem{dalbah2023radarformer}
Y.~Dalbah, J.~Lahoud, and H.~Cholakkal, ``{RadarFormer}: Lightweight and
  accurate real-time radar object detection model,'' in \emph{Image
  Analysis}.\hskip 1em plus 0.5em minus 0.4em\relax Springer, 2023, pp.
  341--358.

\bibitem{palffy2020rtcnet}
A.~Palffy, J.~Dong, J.~F.~P. Kooij, and D.~M. Gavrila, ``{CNN}-based road user
  detection using the {3D} radar cube,'' \emph{IEEE Robotics and Automation
  Letters}, vol.~5, no.~2, pp. 1263--1270, 2020.

\bibitem{zhang2021raddet}
A.~Zhang, F.~E. Nowruzi, and R.~Lagani{\`e}re, ``{RADDet}:
  Range-azimuth-{Doppler} based radar object detection for dynamic road
  users,'' in \emph{2021 18th Conference on Robots and Vision (CRV)}, 2021, pp.
  95--102.

\bibitem{cheng2025transrad}
L.~Cheng and S.~Cao, ``{TransRAD}: Retentive vision transformer for enhanced
  radar object detection,'' \emph{IEEE Transactions on Radar Systems}, vol.~3,
  pp. 303--317, 2025.

\bibitem{zhang2026ptf}
C.~Zhang, Z.~Wang, T.~Fei, C.~Grimm, Y.~Jin, C.~Tebruegge, E.~Warsitz, and
  M.~Gardill, ``Leveraging transformer decoder for automotive radar object
  detection,'' \emph{arXiv preprint arXiv:2601.13386}, 2026.

\bibitem{rebut2022radial}
J.~Rebut, A.~Ouaknine, W.~Malik, and P.~P{\'e}rez, ``Raw high-definition radar
  for multi-task learning,'' in \emph{Proceedings of the IEEE/CVF Conference on
  Computer Vision and Pattern Recognition (CVPR)}, 2022, pp. 17\,021--17\,030.

\bibitem{yang2023adcnet}
B.~Yang, I.~Khatri, M.~Happold, and C.~Chen, ``{ADCNet}: Learning from raw
  radar data via distillation,'' \emph{arXiv preprint arXiv:2303.11420}, 2023.

\bibitem{giroux2023tfftradnet}
J.~Giroux, M.~Bouchard, and R.~Lagani{\`e}re, ``{T-FFTRadNet}: Object detection
  with {Swin} vision transformers from raw {ADC} radar signals,'' in \emph{2023
  IEEE/CVF International Conference on Computer Vision Workshops (ICCVW)},
  2023, pp. 4032--4041.

\bibitem{wu2024sparseradnet}
J.~Wu, M.~Meuter, M.~Schoeler, and M.~Rottmann, ``{SparseRadNet}: Sparse
  perception neural network on subsampled radar data,'' in \emph{European
  Conference on Computer Vision}.\hskip 1em plus 0.5em minus 0.4em\relax
  Springer, 2024, pp. 52--69.

\bibitem{zhuang2025rawadc}
L.~Zhuang, Y.~Yao, T.~Yang, Z.~Wang, and T.~Zhang, ``Boosting {FMCW} radar heat
  map object detection with raw {ADC} data,'' \emph{IEEE Robotics and
  Automation Letters}, vol.~10, no.~11, pp. 12\,087--12\,094, 2025.

\bibitem{fang2025cubedn}
Y.~Fang, F.~Shi, X.~Wei, Q.~Chen, K.~Chetty, and S.~Julier, ``{CubeDN}:
  Real-time drone detection in {3D} space from dual {mmWave} radar cubes,'' in
  \emph{2025 IEEE International Conference on Robotics and Automation (ICRA)},
  2025, pp. 3977--3983.

\bibitem{wei2026maepose}
X.~Wei, Y.~Fang, K.~Chetty, Y.~Cho, and N.~Bianchi-Berthouze, ``{MAEPose}:
  Self-supervised spatiotemporal learning for human pose estimation on mmwave
  video,'' \emph{arXiv preprint arXiv:2605.00242}, 2026.

\bibitem{paek2022kradar}
D.-H. Paek, S.-H. Kong, and K.~T. Wijaya, ``{K-Radar}: {4D} radar object
  detection for autonomous driving in various weather conditions,'' in
  \emph{Advances in Neural Information Processing Systems}, vol.~35.\hskip 1em
  plus 0.5em minus 0.4em\relax Curran Associates, Inc., 2022, pp. 3819--3829.

\bibitem{paek2023enhancedkradar}
------, ``Enhanced {K-Radar}: Optimal density reduction to improve detection
  performance and accessibility of {4D} radar tensor-based object detection,''
  in \emph{2023 IEEE Intelligent Vehicles Symposium (IV)}.\hskip 1em plus 0.5em
  minus 0.4em\relax IEEE, 2023, pp. 1--6.

\bibitem{kong2025rtnhplus}
S.-H. Kong, D.-H. Paek, and S.~Lee, ``{RTNH+}: Enhanced {4D} radar object
  detection network using two-level preprocessing and vertical encoding,''
  \emph{IEEE Transactions on Intelligent Vehicles}, vol.~10, no.~2, pp.
  1427--1440, 2025.

\bibitem{cheng2023centerradarnet}
J.-H. Cheng, S.-Y. Kuan, H.-I. Liu, H.~Latapie, G.~Liu, and J.-N. Hwang,
  ``{CenterRadarNet}: Joint {3D} object detection and tracking framework using
  {4D} {FMCW} radar,'' in \emph{2024 IEEE International Conference on Image
  Processing (ICIP)}, 2024, pp. 998--1004.

\bibitem{madani2022radatron}
S.~Madani, J.~Guan, W.~Ahmed, S.~Gupta, and H.~Hassanieh, ``{Radatron}:
  Accurate detection using multi-resolution cascaded {MIMO} radar,'' in
  \emph{Computer Vision -- ECCV 2022}.\hskip 1em plus 0.5em minus 0.4em\relax
  Springer, 2022, pp. 160--178.

\bibitem{roldan2024radelft}
I.~Roldan, A.~Palffy, J.~F.~P. Kooij, D.~M. Gavrila, F.~Fioranelli, and
  A.~Yarovoy, ``A deep automotive radar detector using the {RaDelft} dataset,''
  \emph{IEEE Transactions on Radar Systems}, vol.~2, pp. 1062--1075, 2024.

\bibitem{sun2025automatic}
B.~Sun, I.~Roldan, and F.~Fioranelli, ``Automatic labelling \& semantic
  segmentation with {4D} radar tensors,'' in \emph{ICASSP 2025 - 2025 IEEE
  International Conference on Acoustics, Speech and Signal Processing
  (ICASSP)}, 2025, pp. 1--5.

\bibitem{ding2024radarocc}
F.~Ding, X.~Wen, Y.~Zhu, Y.~Li, and C.~X. Lu, ``{RadarOcc}: Robust {3D}
  occupancy prediction with {4D} imaging radar,'' \emph{Advances in Neural
  Information Processing Systems}, vol.~37, pp. 101\,589--101\,617, 2024.

\bibitem{major2019vehicle}
B.~Major, D.~Fontijne, A.~Ansari, R.~T. Sukhavasi, R.~Gowaikar, M.~Hamilton,
  S.~Lee, S.~Grzechnik, and S.~Subramanian, ``Vehicle detection with automotive
  radar using deep learning on range-azimuth-{Doppler} tensors,'' in \emph{2019
  IEEE/CVF International Conference on Computer Vision Workshop (ICCVW)}, 2019,
  pp. 924--932.

\bibitem{zheng2023tdradarnet}
R.~Zheng, S.~Sun, H.~Liu, and T.~Wu, ``Deep-neural-network-enabled vehicle
  detection using high-resolution automotive radar imaging,'' \emph{IEEE
  Transactions on Aerospace and Electronic Systems}, vol.~59, no.~5, pp.
  4815--4830, 2023.

\end{thebibliography}

\end{document}